\documentclass[runningheads]{llncs}

\usepackage{eccv} 

\usepackage{eccvabbrv}
\usepackage{graphicx}
\usepackage{booktabs}
\usepackage{pifont}
\usepackage{siunitx}
\usepackage{paralist}
\usepackage{multirow}
\usepackage{tabularray}
\usepackage{wrapfig}
\usepackage[export]{adjustbox}
\usepackage[dvipsnames]{xcolor}
\usepackage[accsupp]{axessibility} 

\usepackage{hyperref} 

\usepackage{orcidlink}

\DeclareGraphicsExtensions{.pdf,.jpg,.png}
\graphicspath{{figs/}}

\newcommand{\secref}[1]{Section~\ref{sec:#1}}
\newcommand{\figref}[1]{Figure~\ref{fig:#1}}
\newcommand{\tabref}[1]{Table~\ref{tbl:#1}}

\newcommand{\siren}{{\sc siren}}

\begin{document}

\title{Probabilistic Image-Driven Traffic \\ Modeling via Remote Sensing}
\titlerunning{Probabilistic Image-Driven Traffic Modeling via Remote Sensing}

\author{Scott Workman\orcidlink{0000-0002-7145-7484} and Armin Hadzic\orcidlink{0000-0002-8816-9366}}
\authorrunning{S.~Workman and A.~Hadzic}
\institute{DZYNE Technologies}

\maketitle

\begin{abstract}

    This work addresses the task of modeling spatiotemporal traffic patterns directly from overhead imagery, which we refer to as image-driven traffic modeling. We extend this line of work and introduce a multi-modal, multi-task transformer-based segmentation architecture that can be used to create dense city-scale traffic models. Our approach includes a geo-temporal positional encoding module for integrating geo-temporal context and a probabilistic objective function for estimating traffic speeds that naturally models temporal variations. We evaluate our method extensively using the Dynamic Traffic Speeds (DTS) benchmark dataset and significantly improve the state-of-the-art. Finally, we introduce the DTS++ dataset to support mobility-related location adaptation experiments.

\end{abstract}

\section{Introduction}

The relationship between humans, the physical environment, and motion guides a number of real-world applications. As Chen et al.~\cite{chen2016promises} note, ``transportation researchers have long sought to develop models to predict how people travel in time and space and seek to understand the factors that affect travel-related choices.'' For example, mobility has been shown to have a strong influence on traffic safety~\cite{wang2013spatio}, public health~\cite{marshall2014community}, housing prices~\cite{chakrabarti2022does}, and more. Recently, with the growth of autonomous driving efforts, there has been an increased focus on using vision-based methods to characterize the physical environment~\cite{geiger2012we,janai2020computer} and how it is traversed~\cite{cui2019multimodal,workman2020dynamic}. 

\begin{figure}
    \centering
    \includegraphics[width=.49\linewidth]{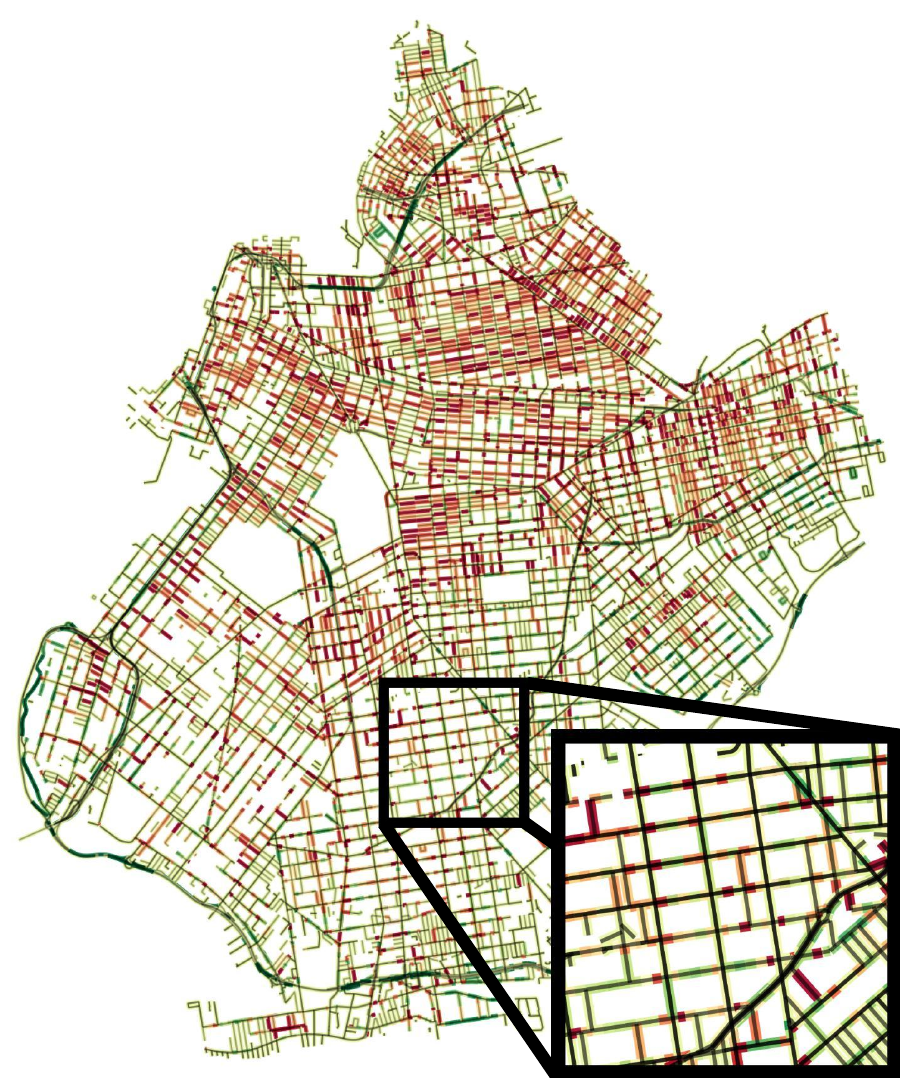}
    \includegraphics[width=.49\linewidth]{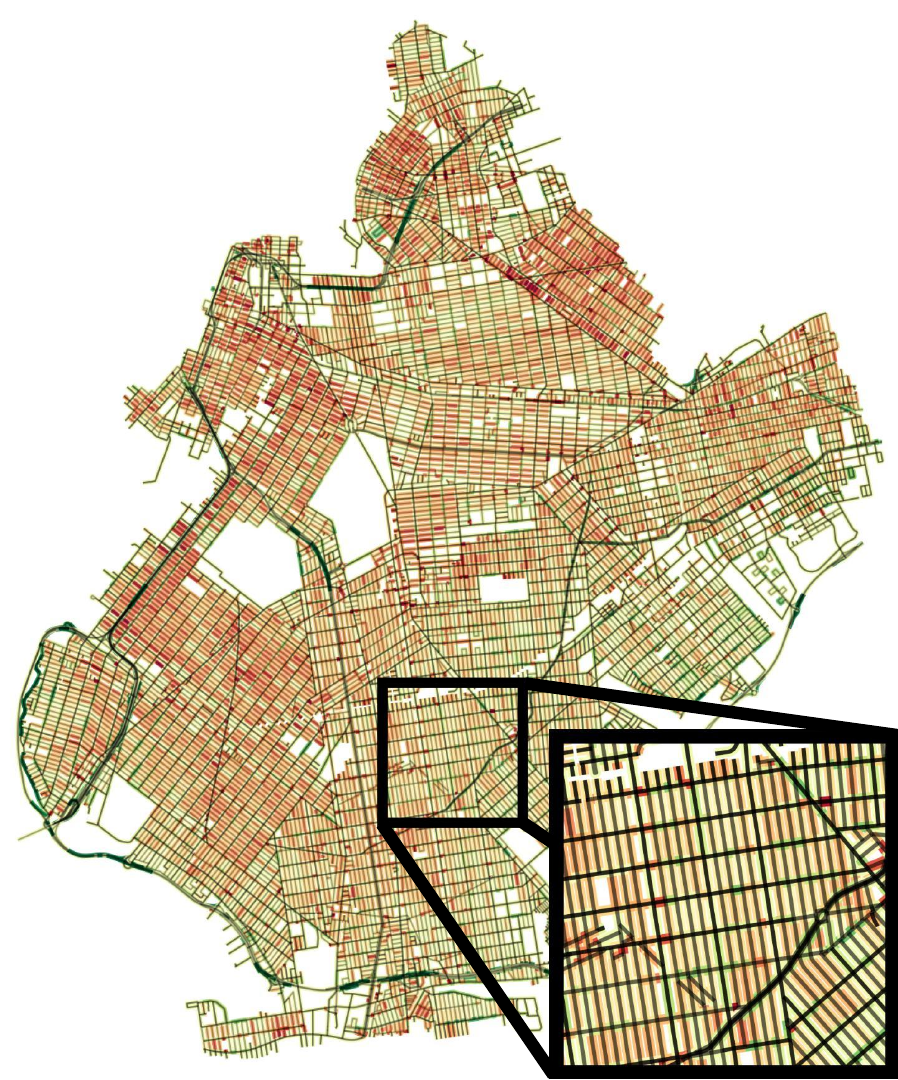}
    \caption{We propose a method for image-driven traffic modeling, which can be used to create dense city-scale traffic models. (left) Historical ground-truth traffic data is often sparse as not all roads are traversed at all times. For example in Brooklyn on Monday at 8am, many roads are missing empirical speed data. (right) Our method can create a dense model of traffic flow at the same time. }
    \label{fig:cartoon}
\end{figure}

A primary challenge that remains is how to scale mobility-related analysis to the size of a city. For example, traffic speed data is predominately collected using fixed-point sensors deployed at static locations on roads~\cite{bickel2007measuring}. Though an increasing amount of traffic speed data is available from alternative sources, such as automatic vehicle location systems, much of this data is still proprietary~\cite{mahajan2022data}. Even in the best case scenario, not every road is traversed at every possible time, resulting in partial, incomplete models of traffic flow (\figref{cartoon}, left).

This work addresses the task of using overhead imagery to directly model spatiotemporal mobility patterns, which we refer to as image-driven traffic modeling. We introduce a multi-modal, multi-task transformer-based segmentation architecture that operates on overhead imagery and can be used to create dense, city-scale models of traffic flow (\figref{cartoon}, right). Our approach integrates geo-temporal context (i.e., geographic location and time metadata) to enable location and time-dependent traffic speed predictions, along with two auxiliary tasks (road segmentation and orientation estimation). These auxiliary tasks provide synergy for traffic speed estimation through multi-task learning, as well as allowing our approach to generalize to locations where the road network is unknown or has changed.

Our approach has several key components. First, we integrate context into our method by introducing a novel geo-temporal positional encoding (GTPE) module. GTPE operates on geo-temporal context through three distinct pathways corresponding to location, time, and space-time features, ultimately producing a positional encoding that captures mobility-context. Second, we propose a probabilistic formulation for estimating traffic speeds that naturally models temporal variation in empirical speed data. Specifically, we capture uncertainty by estimating per-pixel prior distributions over traffic speeds, instead of regressing traffic speeds directly. To support this, we introduce an objective function that explicitly accounts for the number of traffic observations at a given time, during model training. This allows our method access to an implicit form of confidence in the underlying traffic speed averages for a given road segment.

Extensive experiments on a recent benchmark dataset, including ablation studies, demonstrate how our approach achieves superior performance versus baselines. In addition, we extend this dataset to include a new, diverse, city and show how it can be used to support mobility-related location adaptation experiments. We also illustrate several scenarios where our approach could be applied to urban planning applications. Overall, the contributions of this work can be summarized as follows:
\begin{compactitem}
    \item a multi-modal, multi-task architecture for image-driven traffic modeling,
    \item a novel approach for integrating geo-temporal context in the form of a geo-temporal positional encoding module,
    \item a probabilistic formulation for estimating traffic speeds that incorporates knowledge of the number of traffic observations as a form of confidence,
    \item an extensive evaluation of our methods on the Dynamic Traffic Speeds (DTS) benchmark, achieving state-of-the-art results,
    \item and a new dataset, DTS++, to support mobility-related location adaptation experiments.
\end{compactitem}
\section{Related Work}

Traffic modeling is important for many applications including urban planning~\cite{hamilton2005improving} and autonomous driving~\cite{chao2020survey}. These applications ultimately require: 1) descriptions of the underlying environment and 2) knowledge of how environments are traversed. For the former, descriptions of the static environment might include properties such as the location of roads, number of lanes, traffic directions, traffic signs, etc. For the latter, information relating human activity to the underlying infrastructure is necessary, such as time-varying traffic speeds, models of traffic congestion and safety, and other characterizations of driver and pedestrian behavior. Decades of research has focused on understanding related topics in urban mobility.

Numerous learning-based methods have been proposed for relating properties of the environment to how it is traversed. For example, recent work in autonomous driving has explored lane detection from ground-level images~\cite{feng2022rethinking} and how to estimate road layout and vehicle occupancy in top views given a front-view monocular image~\cite{yang2021projecting}. Related to motion, Chen et al.~\cite{chen2021estimating} use traffic cameras to generate mobility statistics from pedestrians and vehicles. Kumar et al.~\cite{kumar2022citywide} propose a method for estimating traffic flow using vehicle-mounted cameras and showcase it in a popular driving simulator. Zhang et al.~\cite{zhang2020spatio} introduce a graph convolutional framework for traffic forecasting that captures spatiotemporal dependencies. Many other vision-based methods have been proposed for estimating vehicle speeds from ground-level imagery; refer to~\cite{fernandez2021vision} for an in-depth overview. 

\begin{figure}
    \centering
    \includegraphics[width=1\linewidth]{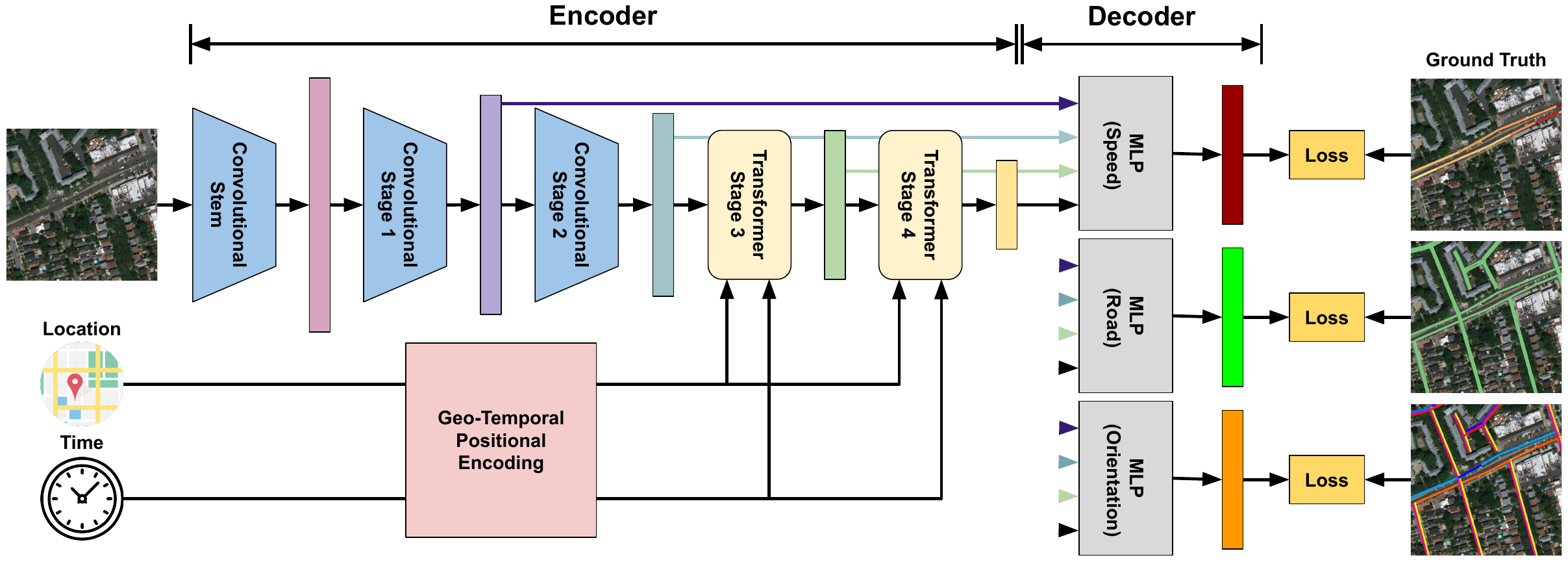}
    \caption{An overview of our architecture for image-driven traffic modeling.}
    \label{fig:architecture}
\end{figure}

Though much progress has been made, a primary challenge that remains is how to build descriptions of the environment at a global scale. For example, traditional approaches for modeling traffic flow assume prior knowledge of the road network and its connectivity~\cite{guo2019attention}, as well as the existence of large quantities of historical traffic data for each road segment~\cite{li2021traffic}. As such, these methods are unable to function in locations where the road network is unknown or where road segments have insufficient historical data. To circumvent issues of scale, overhead imagery has become a useful resource for many tasks due to its dense coverage, high resolution, and ever increasing revisit rates~\cite{albert2017using,ma2019deep}.

For traffic modeling, overhead imagery has been used to automatically generate maps of road networks~\cite{mattyus2017deeproadmapper,zhou2018d}, detect lane boundaries~\cite{homayounfar2019dagmapper}, understand roadway safety~\cite{najjar2017combining}, approximate traffic noise~\cite{eicher2022traffic}, estimate emissions~\cite{mukherjee2021towards}, and many other forms of image-driven mapping~\cite{workman2017unified,workman2017beauty,salem2018soundscape,salem2020learning,vargas2020openstreetmap,workman2022revisiting}. Related to our work, Song et al.~\cite{song2019remote} and Hadzic et al.~\cite{hadzic2020rasternet} show how overhead imagery can be used to understand motion in the form of free-flow traffic speeds (i.e., average speed without adverse conditions). Workman et al.~\cite{workman2020dynamic} introduce the task of dynamic traffic modeling and a new benchmark dataset for traffic speed estimation. We extend this line of work in several ways, including introducing a probabilistic formulation for estimating traffic speeds.
\section{An Architecture for Image-Driven Traffic Modeling}

We address the problem of image-driven traffic modeling using overhead imagery. Our network architecture, depicted in~\figref{architecture}, has three inputs: an overhead image, $S(l)$, a geolocation, $l$, and a time, $t$. It has three outputs, corresponding to our primary task of traffic speed estimation and two auxiliary tasks: road segmentation and orientation estimation. We start with an overview of our multi-task transformer-based segmentation architecture (\secref{segmentation}). Next, we describe our approach for integrating geo-temporal context via a novel geo-temporal positional encoding module (\secref{context}). Then, we describe our proposed probabilistic formulation for estimating traffic speeds (\secref{speeds}). Finally, we detail the loss functions for the auxiliary tasks (\secref{auxiliary}). Please refer to the supplemental material for an in-depth description of architecture design choices.

\subsection{Architecture Overview}
\label{sec:segmentation}

Our segmentation architecture has two primary components: 1) a multi-stage visual encoder for extracting features from an input overhead image, and 2) task-specific decoders which use these features to generate a segmentation output.

\paragraph{Multi-stage Visual Encoder}

Drawing inspiration from CoAtNet~\cite{dai2021coatnet}, which demonstrates that combining convolutional and attention layers can achieve better generalization and capacity, we use a robust multi-stage pipeline for visual feature encoding that integrates both convolutional stages and transformer stages. First, an input image is passed through an initial convolutional stem with three convolutional layers (each with BatchNorm and  ReLU), downsampling the spatial resolution by two. This is followed by two convolutional stages, each of which uses multiple inverted residual blocks in the style of EfficientNet~\cite{hu2018squeeze}. Following the two convolutional stages are two transformer stages inspired by SwinV2~\cite{liu2022swin}. Each stage consists of an overlapping patch embedding~\cite{wang2022pvt} followed by a sequence of multi-head self-attention (MHSA) blocks to allow global information to be propagated throughout the visual features. Each self-attention block uses a learned relative positional encoding~\cite{ke2020rethinking, liu2021swin}.

\paragraph{Task-Specific MLP Decoder}
\label{sec:decoder}

To support semantic segmentation, we extend the multi-stage visual encoder with a task-specific decoder for each prediction task. Taking inspiration from SegFormer~\cite{xie2021segformer}, we use a decoder consisting only of linear layers. Features from each encoder stage are fused via the following process. First, a stage-specific linear layer is used to unify the number of channels across features. The resulting features are then upsampled to a fourth of the resolution of the input image using bilinear interpolation and are then concatenated. This is followed by an additional linear layer that fuses information across features. To produce the segmentation output, a final linear layer operates on the fused feature and generates a task-specific number of outputs. Then, we rescale the output to match the spatial resolution of the input image.

\subsection{Incorporating Geo-Temporal Context}
\label{sec:context}

\begin{figure}[t]
    \centering
    \includegraphics[width=.5\linewidth]{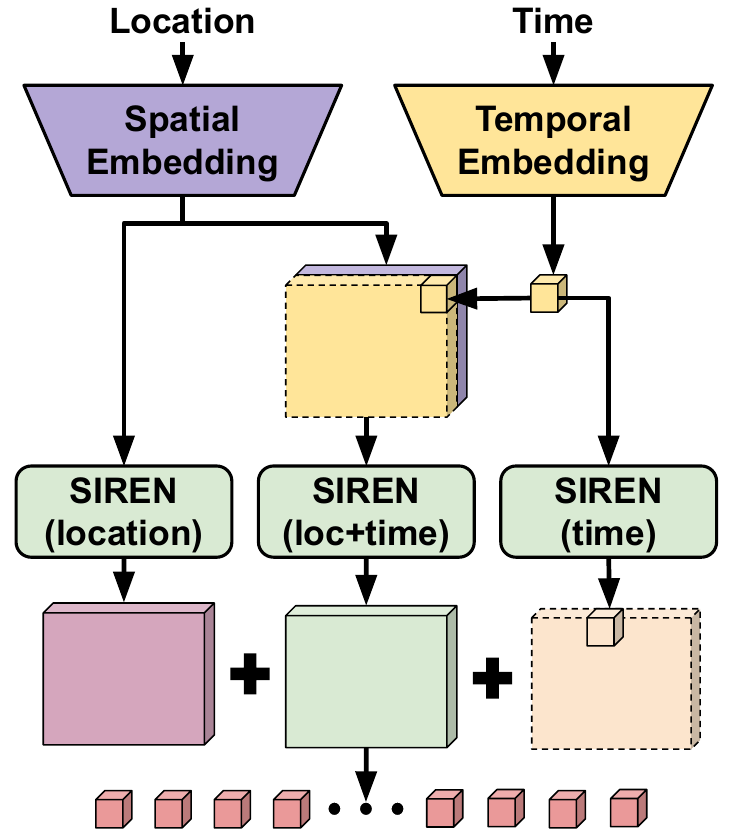}
    \caption{An overview of our proposed geo-temporal positional encoding module.}
    \label{fig:gtpe}
\end{figure}

We infuse our model with an understanding of location and time metadata by introducing a geo-temporal positional encoding module, shown visually in \figref{gtpe}.

\subsubsection{Geo-Temporal Positional Encoding}

We structure the geo-temporal positional encoding (GTPE) module across three pathways corresponding to location, time, and space-time (referred to as loc+time). The location and time pathways reflect that location and time have utility independently, while the space-time pathway allows them to interact. Ultimately, the output of each pathway is added together to form a geo-temporal positional encoding. Next, we detail our choice of context parameterizations and encoding network.

\paragraph{Location} Overhead imagery is unique in that images are typically georeferenced, enabling computation of the world coordinate of each pixel. We leverage this to generate dense location maps in easting ($x$-axis) and northing ($y$-axis) web mercator world coordinates, which are then normalized to a $[-1, 1]$ range using the bounds of the input region.

\paragraph{Time} We represent temporal context using day of week, $d$, and hour of day, $h$, each scaled to $[-1, 1]$. These variables are parameterized using a cyclic representation, similar to Aodha et al.~\cite{mac2019presence}: 
\begin{equation}
    f(d,h) = [ \sin\left(\pi d \right), \cos\left(\pi d \right), \sin\left(\pi h \right), \cos\left(\pi h \right) ].
\end{equation}

\paragraph{Contextual Encoding} We select \siren{}~\cite{sitzmann2020implicit} as the foundation of our encoding network due its ability to extract fine detail from natural data and its representation power for spatial and temporal derivatives. A \siren{} block is composed of a single linear layer followed by a weighted ($W$) sinusoidal activation function ($\sin{(Wa)}$). For each of our contextual features, $a$, we pass the parameterized inputs through an encoding network made up of three \siren{} blocks, each with a hidden dimension of 64 (128 for loc+time), followed by a final linear layer which produces a 64-dimensional embedding. When fusing time with location, we replicate the spatial dimensions to match.

\subsubsection{Fusing with Visual Features} Similar to a traditional positional encoding, we merge the GTPE output with the visual features in the transformer stages of our multi-modal fusion architecture. At each stage, we linearly reproject the GTPE output to match the stage-specific embedding size. This is followed by bilinear interpolation to scale the spatial dimensions to match the spatial resolution at that stage. Finally, we treat the output as a positional encoding and add it to the corresponding visual tokens to infuse the network with geo-temporal context. While our primary task is mobility analysis, GTPE is applicable to other problem domains (see supplemental material).

\subsection{Probabilistic Traffic Speed Estimation}
\label{sec:speeds}

We propose a probabilistic approach for traffic speed estimation that naturally models variations in likely traffic speeds, along with an objective function that explicitly accounts for the number of traffic observations (i.e., a form of confidence in the underlying traffic speed averages for a given road segment). For this we use the Student's t-distribution, a symmetric distribution similar to the normal distribution that is used when estimating the mean of a normally distributed population in scenarios where: 1) there are a small number of observations and 2) the standard deviation of the population is unknown. In other words, a Student's t-distribution relates the distribution of the sample mean to the true mean. 

For our purposes, we use the generalized form of the Student's t-distribution denoted as $t_\nu(\mu, \sigma)$, where $\mu$ is a location (shift) parameter, $\sigma > 0$ is a scale parameter, and $\nu > 0$ is a shape parameter (often referred to as degrees of freedom). The probability density function for this form of the Student's t-distribution is given by:
\begin{equation}
    p(x|\nu,\mu,\sigma) = \frac{\Gamma(\frac{\nu+1}{2})}{\Gamma(\frac{\nu}{2})\sqrt{\nu\pi}\sigma}\left(1 + \frac{1}{\nu}\left[\frac{x - \mu}{\sigma}\right]^2\right)^{-\frac{\nu+1}{2}},
    \label{pdf}
\end{equation}
where $\Gamma$ is the gamma function,
\begin{equation}
    \Gamma(\alpha) = \int_{0}^{\infty} x^{\alpha-1}e^{-x}\,dx.
\end{equation}

Instead of regressing traffic speeds directly, we capture uncertainty by estimating per-pixel prior distributions over traffic speeds. Given an overhead image and geo-temporal context as input, the output of the traffic speed decoder is per-pixel estimates of the shift-scale parameters of the prior distribution (i.e., $\mu$ and $\sigma^2$). A softplus activation is used to ensure these outputs are positive. During model training, we combine the estimated shift-scale parameters with the true count of traffic observations for each road segment, as the shape parameter $\nu$, to form a Student's t distribution.

To optimize our approach, we minimize the negative log likelihood of the resulting distributions, treating ground-truth traffic speeds on a given road segment as samples from the corresponding Student's t-distribution. Specifically, given an observed traffic speed, $y$, we compute the likelihood under the distribution (\ref{pdf}). Our objective function then becomes:
\begin{equation}
    \mathcal{L}_{speed} = - \log G(S(l),l,t;\Theta)(y),
\end{equation}
where $G$ represents our proposed approach which takes as input an overhead image $S$, geolocation $l$, and time $t$, and outputs prior distributions over traffic speeds, and $\Theta$ are the weights of the network, which we optimize. At inference, we use the estimated shift parameter of our output distributions, $\mu$, as our prediction for traffic speed. 

\paragraph{Region Aggregation} We assume ground-truth traffic speeds are provided as averages over road segments (i.e., in aggregate form). However, our network generates dense, full-resolution predictions. We use a variant of the region aggregation layer~\cite{jacobs2018weakly} to facilitate comparison to the ground-truth values. This process averages predictions across a given road segment to produce a single aggregated estimate. In our case, we aggregate the estimated shift and scale parameters for each road segment before forming a per-road-segment Student's t-distribution.

\subsection{Auxiliary Tasks}
\label{sec:auxiliary}

For the auxiliary tasks of road segmentation and orientation estimation, we create a task-specific decoder as described in \secref{decoder}. For road segmentation, we follow recent work and formulate this as a binary segmentation task. As our objective function, we use a combined loss that incorporates binary cross entropy and the Dice loss~\cite{zhou2018d} ($\mathcal{L}_{bce} + (1 -\mathcal{L}_{dice})$). For orientation estimation, we treat this as a classification task over $K$ angular bins, and use standard cross entropy as the objective function. The cumulative objective function for primary and auxiliary tasks becomes:
\begin{equation}
    \mathcal{L} = \mathcal{L}_{speed} + \mathcal{L}_{road} + \mathcal{L}_{orientation}.
\end{equation}

\subsection{Implementation Details}

Our methods are implemented using PyTorch~\cite{paszke2019pytorch} and PyTorch Lightning~\cite{falcon2019pytorch} and optimized using Adam~\cite{kingma2014adam} ($\lambda = 1e^{-4}$). We simultaneously optimize the entire network for all tasks (loss terms weighted equally) and train for 50 epochs. Model selection is performed using a validation set. For each transformer stage, we use eight attention heads in all instances of MHSA. The expansion rate for each inverted residual block is 4. The expansion/shrink rate for the squeeze-and-excitation layers is 0.25. For road segmentation, we buffer road geometries assuming a two meter half width. For orientation estimation, we use $K=16$ angular bins. We train on full size images and traffic speeds are represented in kilometers per hour.
\section{Experiments}

We evaluate our approach for the task of traffic speed estimation through a variety of experiments.

\subsection{Dynamic Traffic Speeds Dataset}

\begin{figure}[t]
    \centering
    \setlength\tabcolsep{1pt}
    \begin{tabular}{ccc|cccc}
        Image & Roads & Orientation & Mon. 8am & Mon. 8pm & Sat. 8am & Sat. 8pm \\
        \includegraphics[width=.137\linewidth]{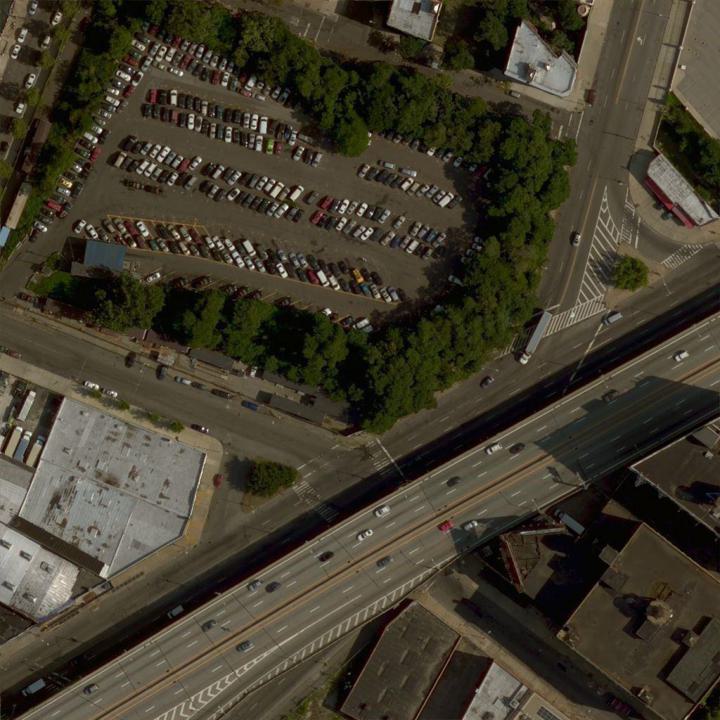} &
        \includegraphics[width=.137\linewidth]{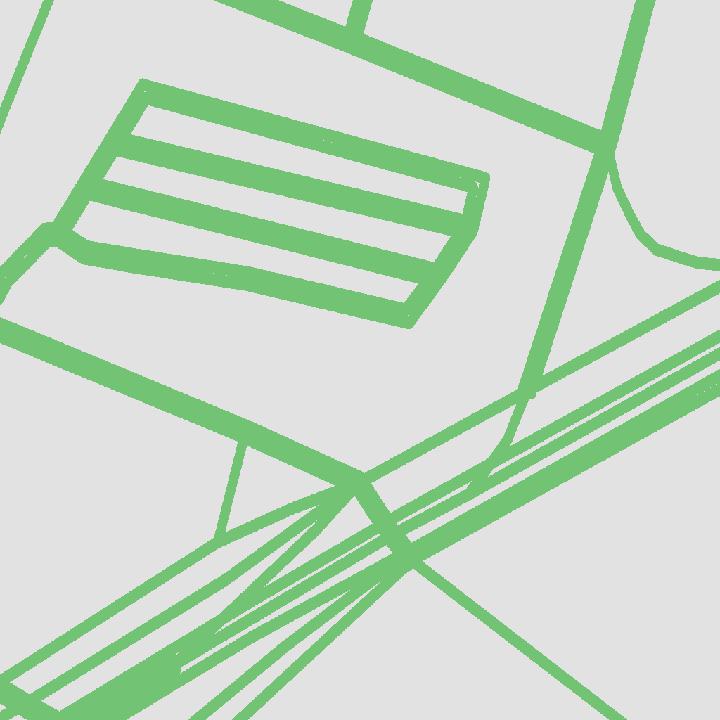} &
        \includegraphics[width=.137\linewidth]{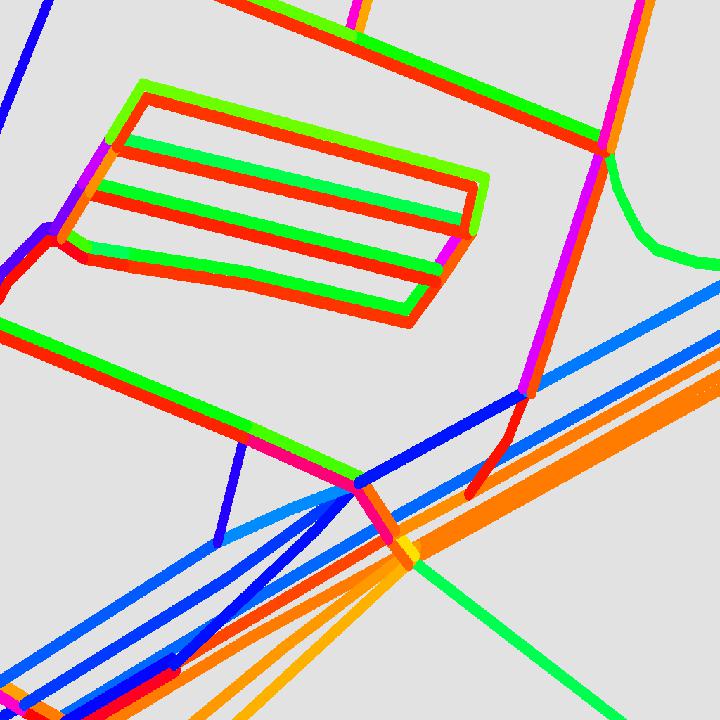} &
        \includegraphics[width=.137\linewidth]{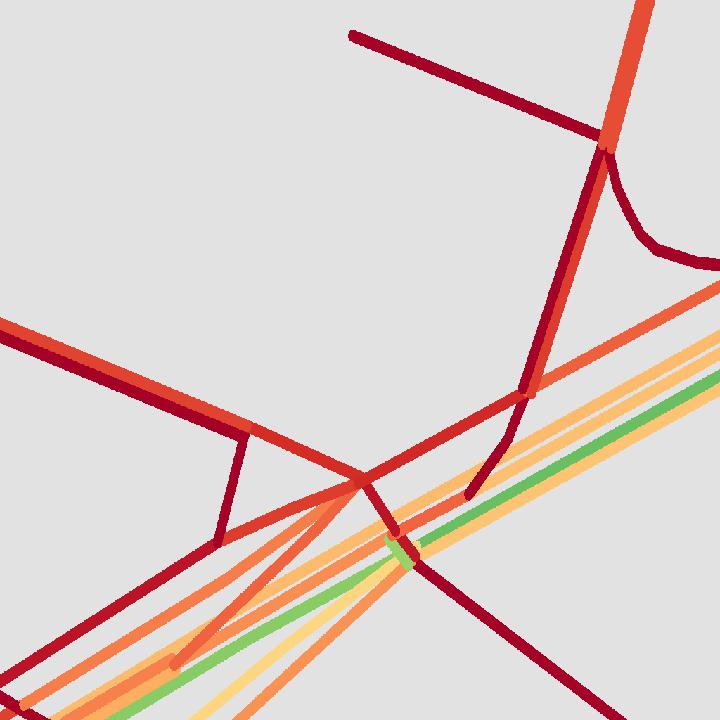} &
        \includegraphics[width=.137\linewidth]{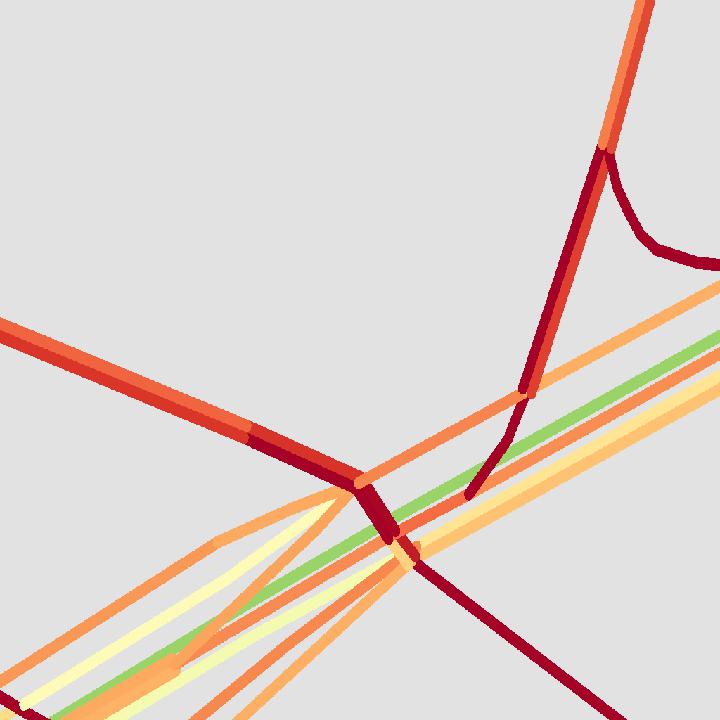} &
        \includegraphics[width=.137\linewidth]{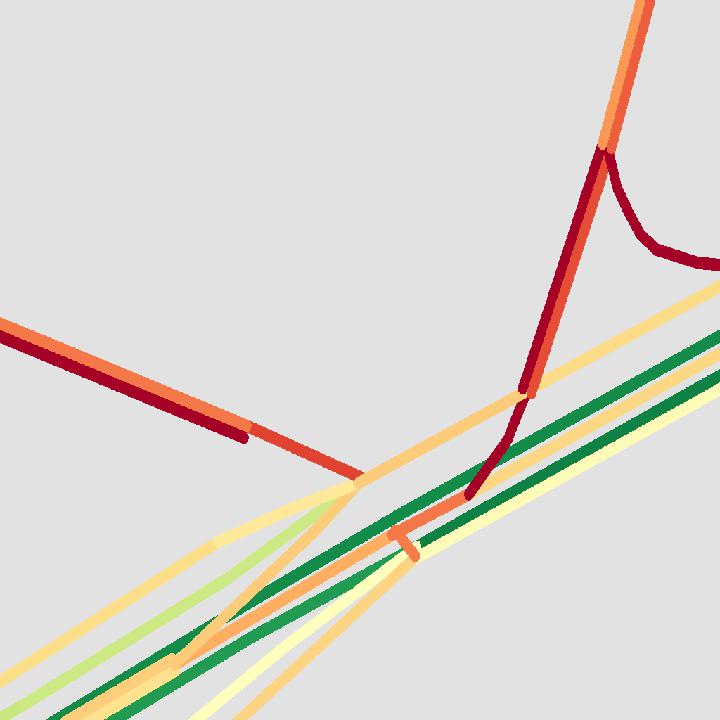} &
        \includegraphics[width=.137\linewidth]{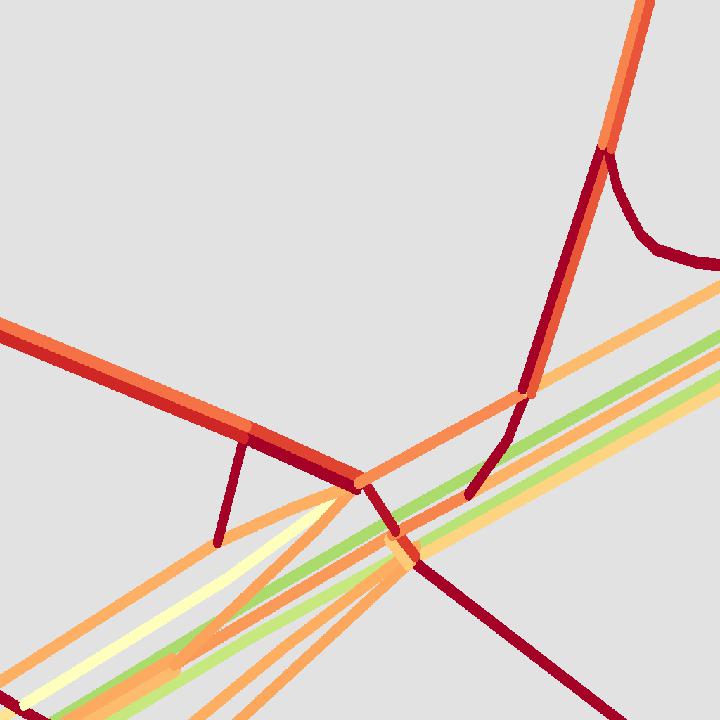} \\
    \end{tabular}
    
    \caption{Example images from the Dynamic Traffic Speeds (DTS) dataset and corresponding labels. The right four labels depict available historical traffic speeds at different times, where green (red) corresponds to faster (slower) speeds.}

  \label{fig:dataset}
\end{figure}

We train and evaluate our method using the recently introduced Dynamic Traffic Speeds (DTS) dataset~\cite{workman2020dynamic}, a fine-grained road understanding benchmark. DTS relates overhead imagery and road metadata with a year of historical traffic speeds for New York City, NY. The traffic speed data originates from Uber Movement Speeds~\cite{movement}, a collection of aggregated speed data at the road segment level, with hourly frequency, from Uber rideshare trips. The dataset contains \num[group-separator={,}]{11902} non-overlapping overhead images ($1024 \times 1024$) at approximately $0.3$ meters per-pixel, with associated road segment information, and corresponding historical traffic data. \figref{dataset} visualizes example data from DTS. We use the original data splits, consisting of 85\% training, 5\% validation, and 10\% testing. Similar to~\cite{workman2020dynamic}, we dynamically generate speed masks during training.

\paragraph{Evaluation Metrics} We report three metrics for traffic speed (km/h) estimation: root-mean-square error (RMSE), mean absolute error (MAE), and the coefficient of determination ($R^2$), a proportion which describes how well variations in the observed values can be explained by the model. When computing evaluation metrics, we apply region aggregation to average predictions along each road segment to enable comparison to the ground truth. 

\subsection{Traffic Speed Estimation}

As each test image is associated with time-varying historical traffic data, we consider two strategies for selecting a time for evaluation. For our first experiment, we sample a time for each test image from the set of observed traffic data across all contained road segments, and use this time to generate a ground-truth speed mask. We refer to this as a micro strategy, as metrics are computed globally across time. \tabref{evaluation_micro} shows the results of this study. Across all metrics, our method significantly outperforms the prior state-of-the-art.

For the second experiment, we employ a macro strategy that considers a specific set of times (Monday \& Saturday, with hours 12am, 4am, 8am, 12pm, 5pm, and 8pm). For each time, we select all images in the test set that  contain a road segment with observed traffic speed data at that time. Metrics are then averaged across time such that all times are weighted equally. \tabref{evaluation_macro} shows the results of this study for a subset of times, with the overall average performance shown in the bottom row. As before, our method significantly outperforms prior work independent of the day of the week or hour of day.

\newlength{\defaultintextsep}
\setlength{\defaultintextsep}{\intextsep}
\setlength{\intextsep}{3pt}
\setlength{\columnsep}{12pt}
\begin{wraptable}{r}{8.5cm}
    \parbox{1\linewidth}{
          \centering
          \caption{Micro evaluation of traffic speed estimation.}
          
          \robustify\bfseries
          \sisetup{text-series-to-math}
          \setlength\tabcolsep{2pt}
          
          \begin{tabular}{@{}lcSSS@{}}
            \toprule
            Method & Loss & \multicolumn{1}{c}{RMSE $\downarrow$} & \multicolumn{1}{c}{MAE $\downarrow$} & \multicolumn{1}{c}{$R^2\uparrow$}  \\
            \bottomrule
            \cite{workman2020dynamic} & Pseudo-Huber & 10.64 & 8.09 & 0.46 \\
            Ours                      & Pseudo-Huber & 10.02 & 7.34 & 0.52 \\
            Ours                      & Student's t  & \bfseries 8.84 & \bfseries 6.64 & \bfseries 0.63 \\
            \bottomrule
          \end{tabular}
          
          \label{tbl:evaluation_micro}
    }
    \hfill
    \parbox{1\linewidth}{
          \centering
          \caption{Macro evaluation of traffic speed estimation.}
          
          \robustify\bfseries
          \sisetup{text-series-to-math}
          \setlength\tabcolsep{2pt}
          
          \begin{tabular}{@{}lSSSSSS@{}}
            \toprule
            & \multicolumn{3}{c}{Workman et al.~\cite{workman2020dynamic}} & \multicolumn{3}{c}{Ours} \\
            & {RMSE} & {MAE} & {$R^2$} & {RMSE} & {MAE} & {$R^2$} \\
            \midrule
            Mon (4am) &  13.20 & 9.74 & 0.63 & 10.95 & 7.85 & 0.75 \\
            Mon (12pm) & 10.40 & 7.86 & 0.52 &  8.78 & 6.56 & 0.66 \\
            Sat (5pm) &  10.35 & 7.96 & 0.43 &  8.68 & 6.60 & 0.60 \\
            Sat (8pm) &  10.26 & 7.78 & 0.46 &  8.47 & 6.34 & 0.63 \\
            \midrule
            Overall   &  11.12 & 8.34 & 0.49 & \bfseries 9.15 & \bfseries 6.76 & \bfseries 0.65 \\
            \bottomrule
          \end{tabular}
          
          \label{tbl:evaluation_macro}
    }
\end{wraptable}

\begin{figure}
    \centering
    
    \setlength\tabcolsep{1pt}
    
    \begin{tabular}{ccccccc}        
        \raisebox{.04\height}{\rotatebox{90}{\scriptsize Ground Truth}} &
        \includegraphics[width=.155\linewidth]{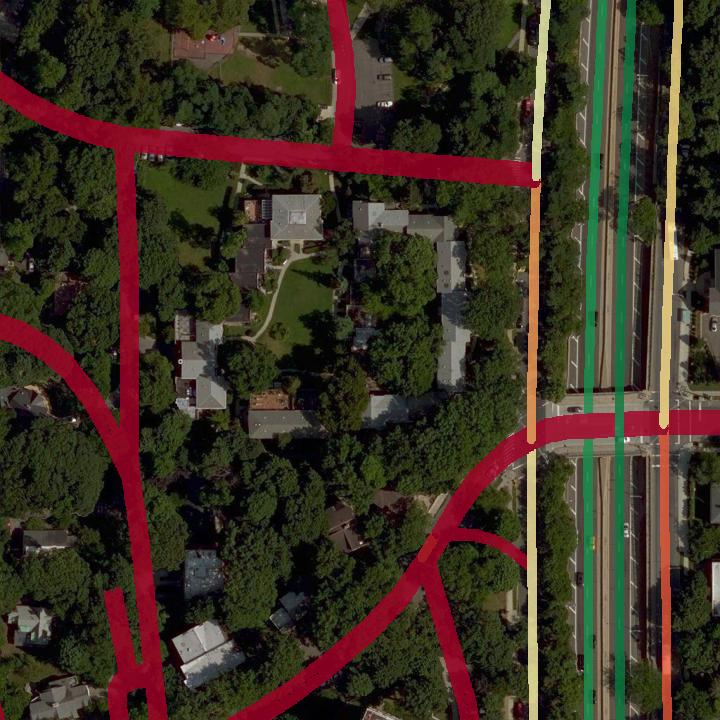} &
        \includegraphics[width=.155\linewidth]{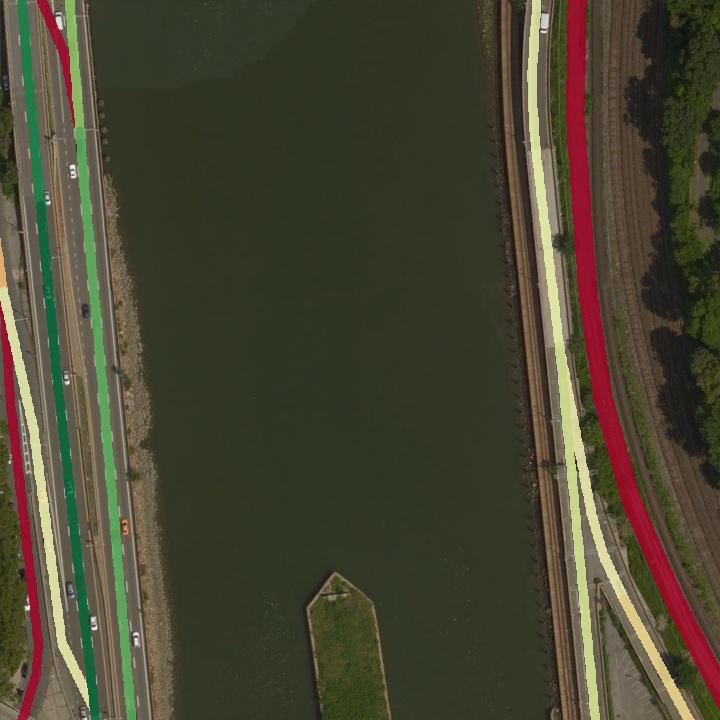} &
        \includegraphics[width=.155\linewidth]{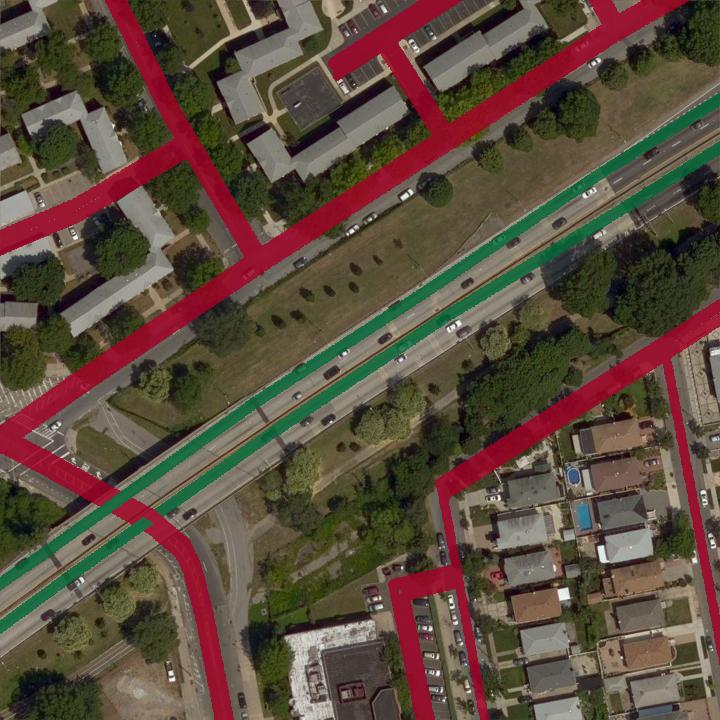} &
        \includegraphics[width=.155\linewidth]{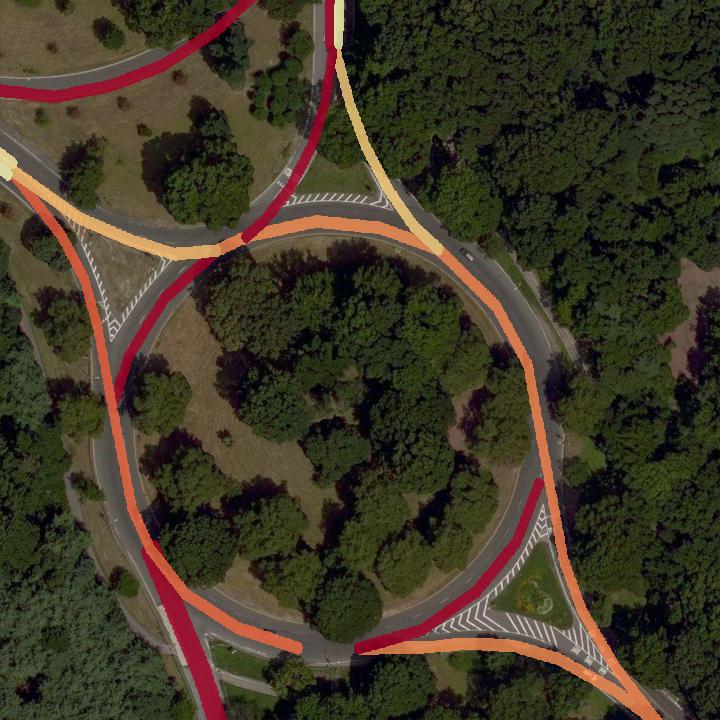} &
        \includegraphics[width=.155\linewidth]{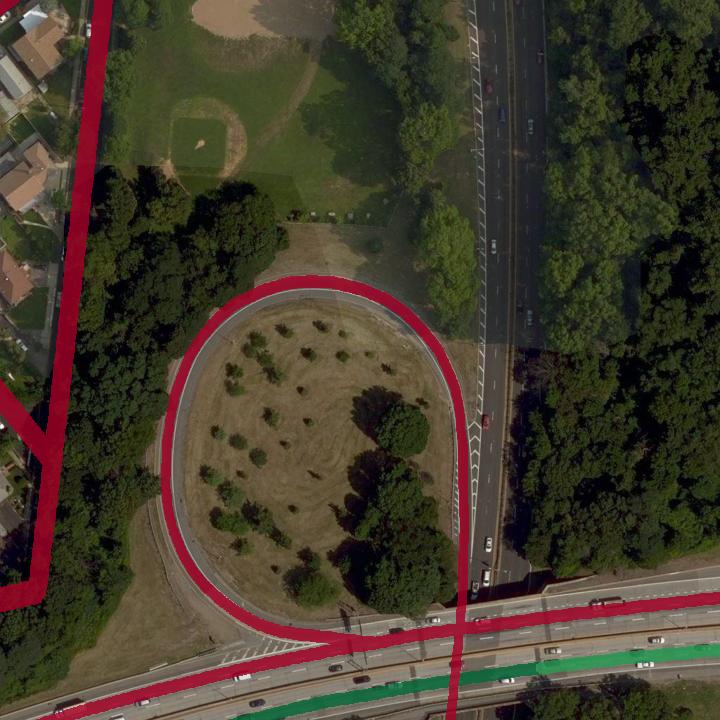} &
        \includegraphics[width=.155\linewidth]{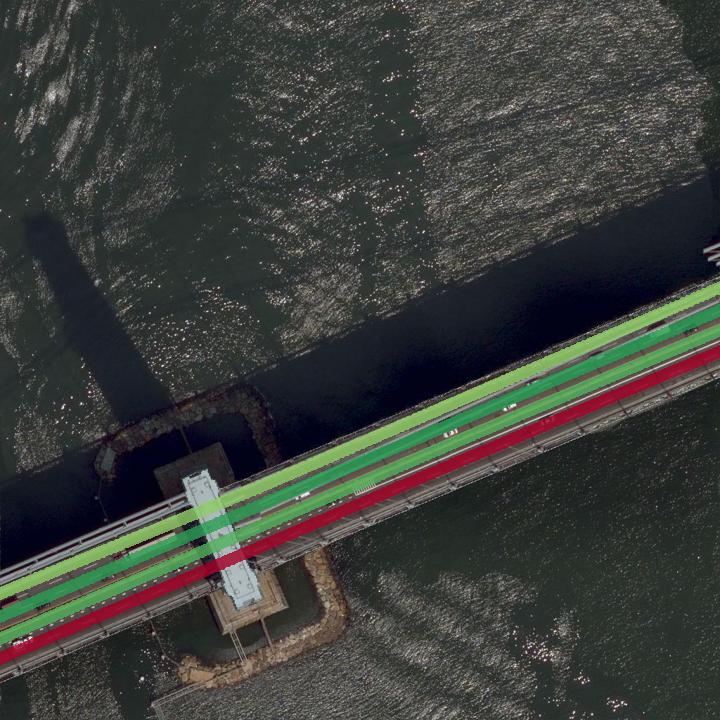} \\
        
        \raisebox{.1\height}{\rotatebox{90}{\scriptsize Pred. (Agg.)}} &
        \includegraphics[width=.155\linewidth]{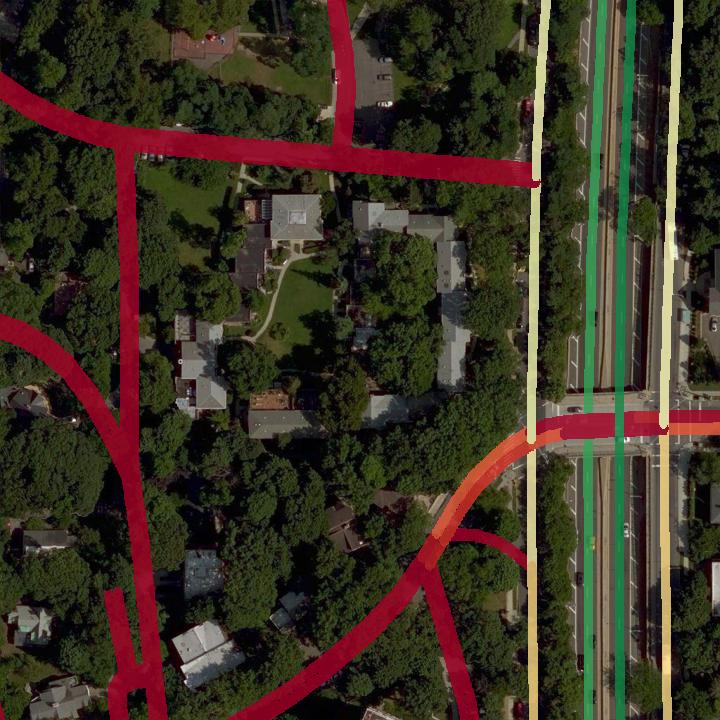} &
        \includegraphics[width=.155\linewidth]{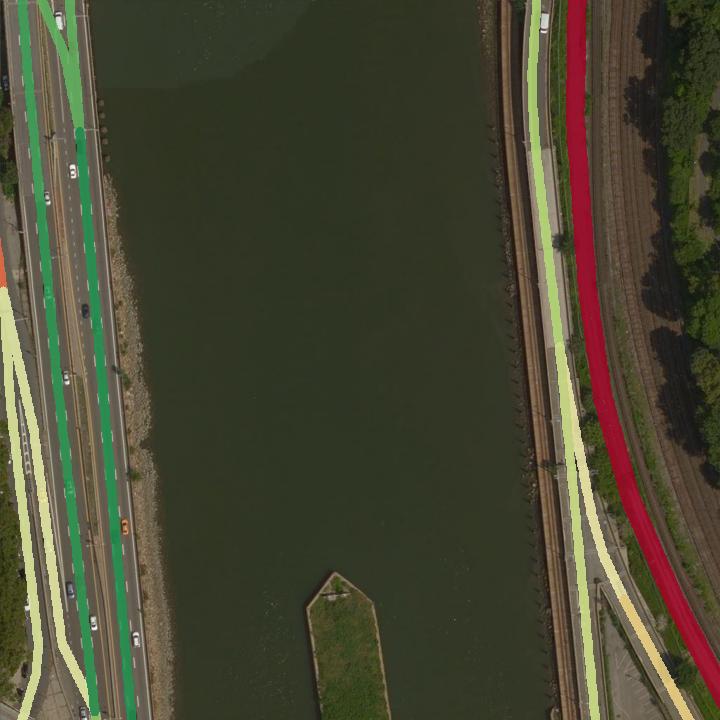} &
        \includegraphics[width=.155\linewidth]{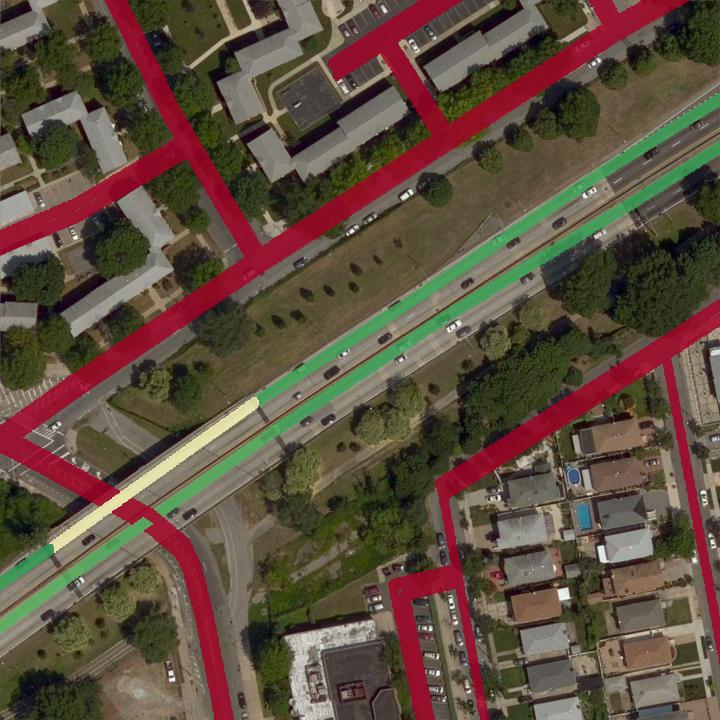} &
        \includegraphics[width=.155\linewidth]{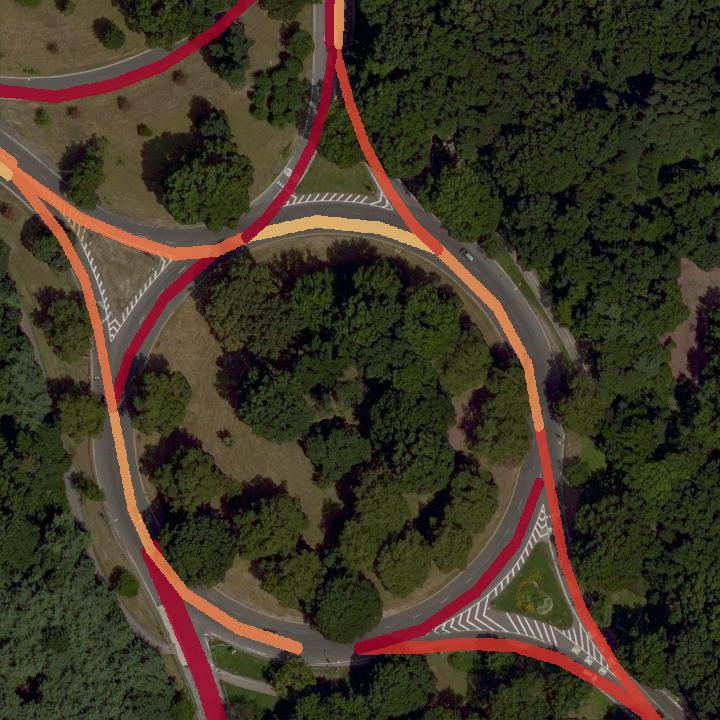} &
        \includegraphics[width=.155\linewidth]{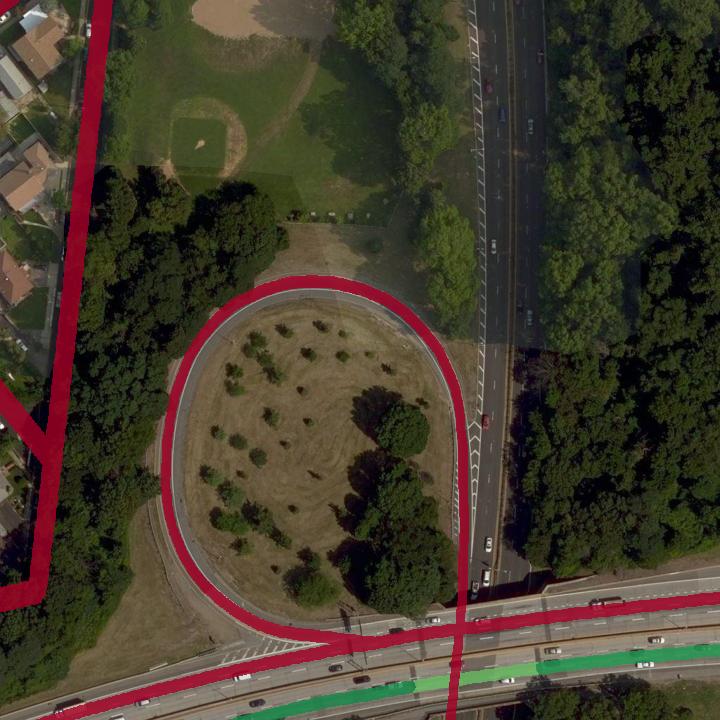} &
        \includegraphics[width=.155\linewidth]{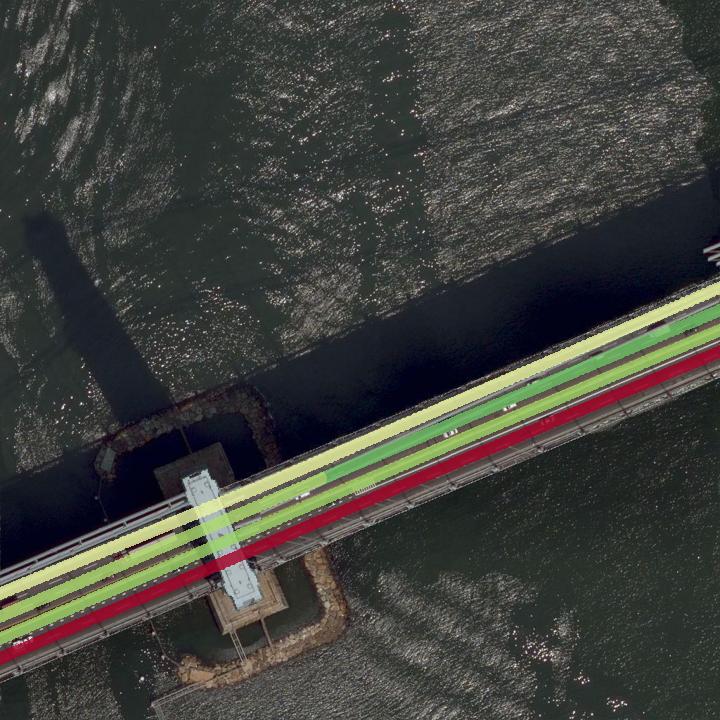} \\
    
        \raisebox{.05\height}{\rotatebox{90}{\scriptsize Pred. (Dense)}} &
        \includegraphics[width=.155\linewidth]{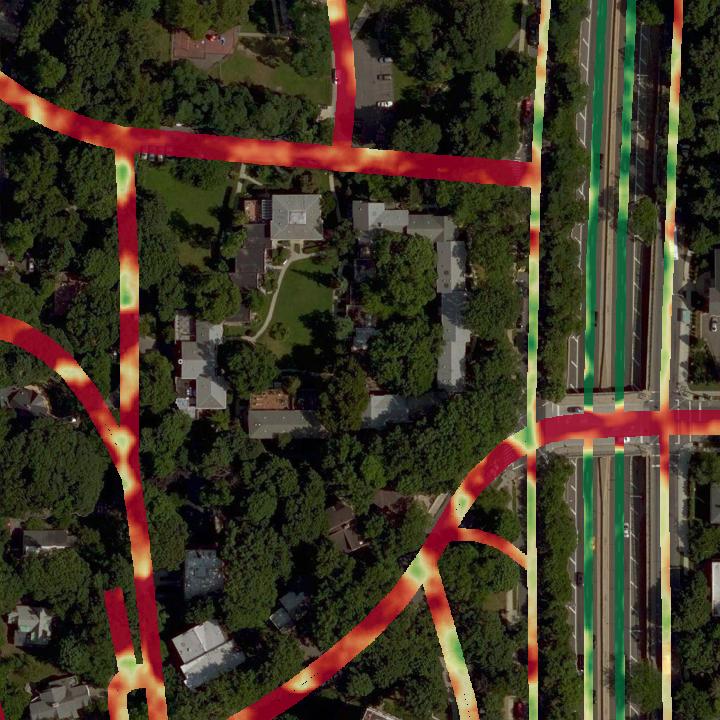} &
        \includegraphics[width=.155\linewidth]{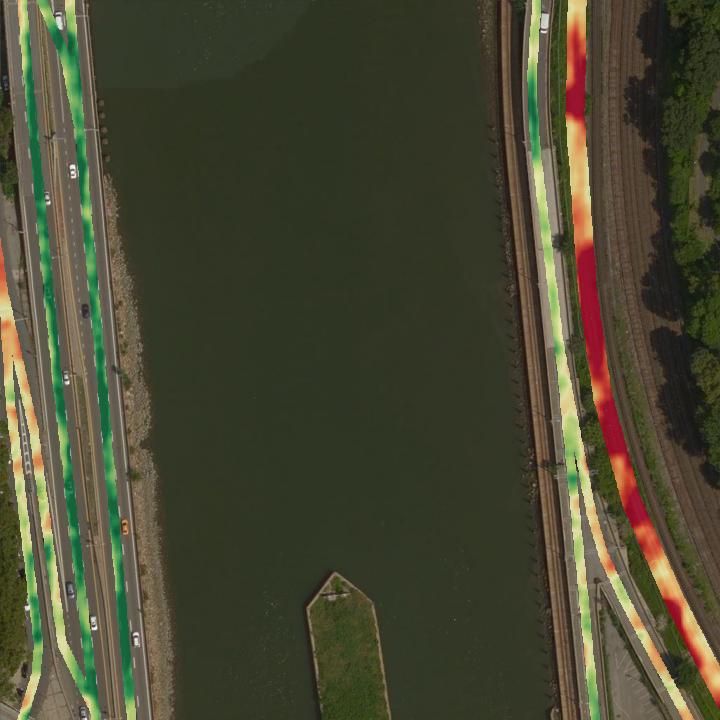} &
        \includegraphics[width=.155\linewidth]{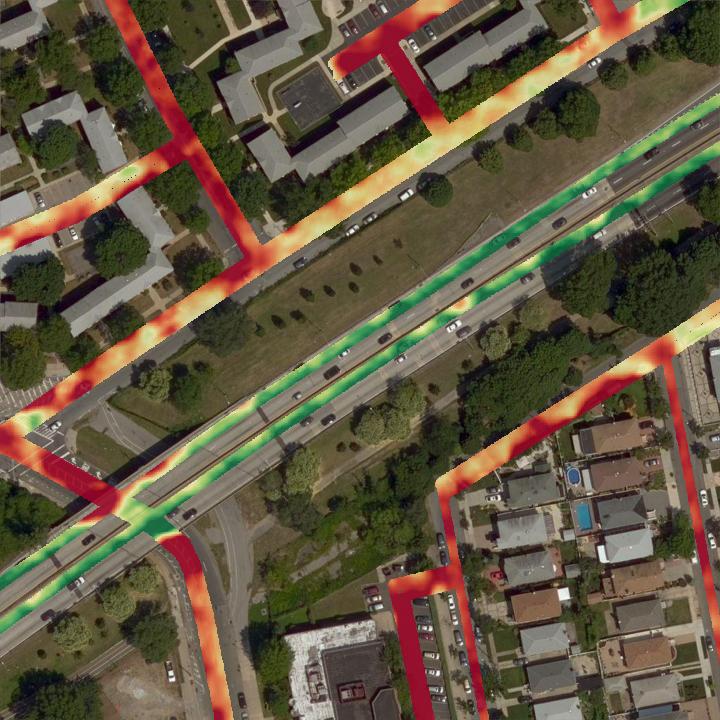} &
        \includegraphics[width=.155\linewidth]{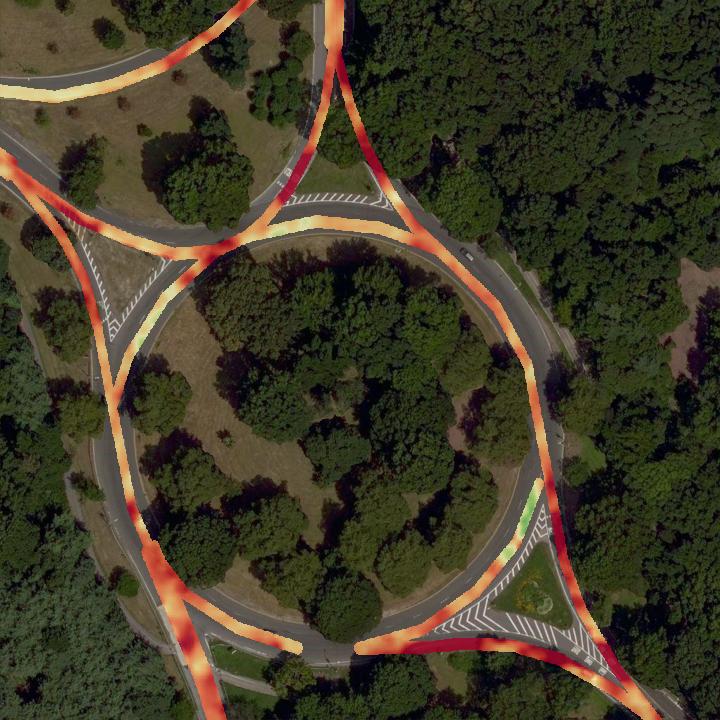} &
        \includegraphics[width=.155\linewidth]{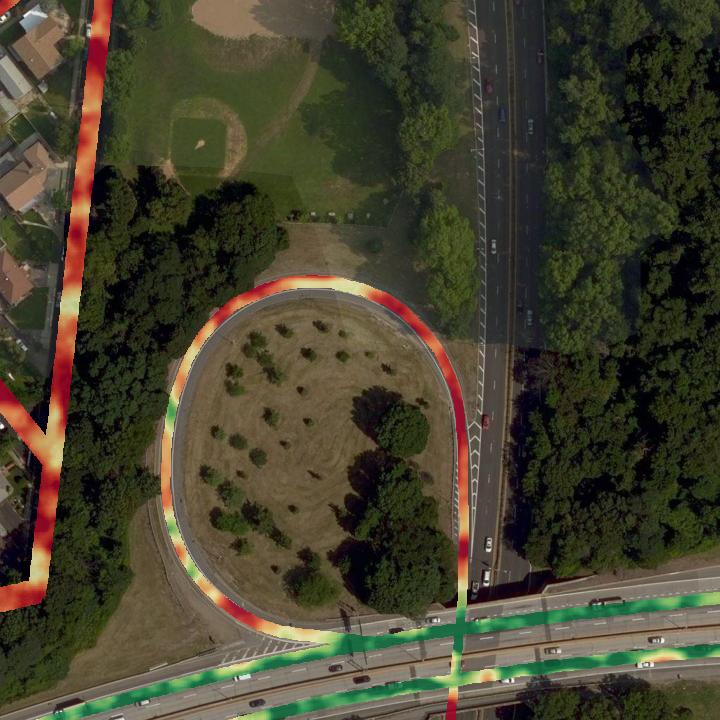} &
        \includegraphics[width=.155\linewidth]{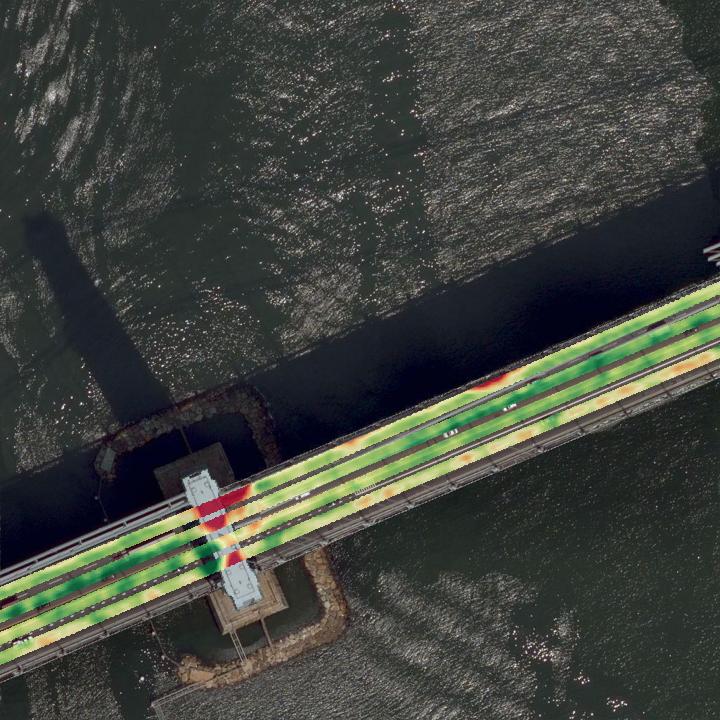} \\
    \end{tabular}
    
    \caption{Qualitative results for traffic speed estimation (Monday, 8am). (top) Ground-truth traffic speeds are provided as aggregates for individual road segments. For visualization, we replicate the ground-truth speed across the entire road segment. (middle) Predictions from our approach, after applying region aggregation. (bottom) Our results without region aggregation capture nuances of traffic flow, such as slowing down around curves. Green (red) corresponds to fast (slow) traffic speeds.}
    \label{fig:qualitative}
\end{figure}

\figref{qualitative} shows qualitative results from our method. The top row shows ground-truth traffic speeds, which are provided as aggregates across road segments. The middle row shows the results of our method, after applying region aggregation to individual road segments. The bottom row shows the dense output of our approach, without region aggregation (per-pixel traffic speeds). As observed, our approach is able to capture nuances of traffic flow. For example, that roundabouts typically have faster traffic in straightaways but slower traffic on on-ramps and exits.

\setlength{\intextsep}{13pt}

\subsection{Ablation Study}
\label{sec:ablation}

We conduct an extensive ablation study to validate components of our approach. First, we consider the choice of objective function. \tabref{evaluation_micro} shows the results of this experiment. We compare our probabilistic approach (Student's t) versus a variant of our method which directly regresses traffic speeds using the Pseudo-Huber loss. As observed, our probabilistic approach significantly improves performance relative to this baseline. The choice of baseline objective function is consistent with that used in the prior state-of-the-art~\cite{workman2020dynamic}, highlighting that our probabilistic formulation directly leads to an increase in performance.

Next, we evaluate the impact of geo-temporal context on our traffic speed predictions. \tabref{ablation_context} shows the results of this experiment. First, an image-only variant of our approach outperforms a metadata-only variant ({\em loc, time}), demonstrating the importance of visual features for this task. Further, adding both location and time context leads to additional performance gains, with time context being superior to location context for predicting time-varying traffic speeds. Finally, like-for-like comparisons with prior work show the superior performance of our approach in each setting.

\setlength{\intextsep}{\defaultintextsep}
\begin{table}[t]
  \centering
  \caption{Ablation study evaluating the impact of location and time context.}

  \setlength\tabcolsep{6pt}
          
  \begin{tabular}{@{}llSSS@{}}
    \toprule
    {Method} & {Context} & {RMSE} & {MAE} & {$R^2$} \\
    \bottomrule
    \multirow{4}{*}{\cite{workman2020dynamic}} 
    & {\em loc, time}        & 13.13 & 9.60 & 0.18 \\
    & {\em image}            & 11.33 & 8.68 & 0.39 \\
    & {\em image, loc}       & 10.94 & 8.43 & 0.43 \\
    & {\em image, time}      & 10.67 & 8.11 & 0.46 \\
    & {\em image, loc, time} & 10.64 & 8.09 & 0.46 \\
    \midrule
    \multirow{4}{*}{Ours} 
    & {\em loc, time}        & 12.16 & 9.20 & 0.30 \\
    & {\em image}            & 10.28 & 7.70 & 0.50 \\
    & {\em image, loc}       &  9.65 & 7.26 & 0.56 \\
    & {\em image, time}      &  9.10 & 6.75 & 0.61 \\
    & {\em image, loc, time} &  8.84 & 6.64 & 0.63 \\
    \bottomrule
  \end{tabular}
  
  \label{tbl:ablation_context}
\end{table}

\begin{table}[b]
  \centering
  \caption{Ablation study comparing different strategies for integrating geo-temporal context.}

  \setlength\tabcolsep{6pt}
  
  \begin{tabular}{@{}cccccSSS@{}}
    \toprule
    {\multirow{2}{*}{Loc. Rep.}} & \multicolumn{2}{c}{Context Enc.} & \multicolumn{2}{c}{Context Fusion} & {\multirow{2}{*}{RMSE}} & {\multirow{2}{*}{MAE}} & {\multirow{2}{*}{$R^2$}}  \\
    & {\em loc} & {\em time} & {\em loc} & {\em time} & & &  \\
    \bottomrule
    Center & MLP  & MLP  & Concat & Concat & 10.64 & 8.09 & 0.46 \\
    Center & MLP  & MLP  & Token  & Token  &  9.13 & 6.87 & 0.61 \\
    Dense  & CNN  & MLP  & PE     & Token  &  9.02 & 6.71 & 0.61 \\
    Dense  & GTPE & GTPE & PE     & PE     &  8.84 & 6.64 & 0.63 \\
    \bottomrule
  \end{tabular}
  
  \label{tbl:ablation_integration}
\end{table}

We also perform an ablation study comparing strategies for integrating geo-temporal context. The results are shown in \tabref{ablation_integration}. For this experiment, we vary the location representation, context encoding, and context fusion method. The results show that using a per-pixel representation (Dense) for location is superior to the center coordinate (Center) as in prior work~\cite{workman2020dynamic}. Additionally, encoding location and time context using GTPE leads to better performance as compared to an MLP or CNN encoder. For fusing context features with visual features, treating location and time as a positional encoding (PE), as in GTPE, performs better than channel-wise concatenating (Concat) these features in the decoder, or adding context features as an additional token (Token) in each transformer stage (e.g., \cite{diao2022metaformer}). In summary, GTPE outperforms other strategies for integrating geo-temporal context, which are representative of the prior state-of-the-art.

\setlength{\intextsep}{3pt}
\begin{wraptable}{r}{6.1cm}
  \centering
  \caption{Ablation study on the impact of multi-task learning.}

  \setlength\tabcolsep{2pt}
  
  \begin{tabular}{@{}ccSSS@{}}
    \toprule
    Road & Orientation & \multicolumn{1}{c}{RMSE} & \multicolumn{1}{c}{MAE} & \multicolumn{1}{c}{$R^2$}  \\
    \bottomrule
    \ding{55} & \ding{55} & 9.18 & 6.86 & 0.60 \\
    \ding{51} & \ding{55} & 9.03 & 6.70 & 0.61 \\
    \ding{55} & \ding{51} & 8.92 & 6.67 & 0.62 \\
    \ding{51} & \ding{51} & 8.84 & 6.64 & 0.63 \\
    \bottomrule
  \end{tabular}
  
  \label{tbl:ablation_task}
\end{wraptable}

Finally, we compare our approach, which includes the auxiliary tasks of road segmentation and orientation estimation, to variants of our approach that consider subsets of the tasks (\tabref{ablation_task}). Results show that including the auxiliary tasks has a positive impact on performance. This matches recent work in multi-task learning that finds sharing information across tasks can lead to performance improvements when the tasks are synergistic~\cite{crawshaw2020multi}.

\subsection{DTS++: A Dataset for Location Adaptation}

Location adaptation, an analog to domain adaptation, aims to address the problem of adapting a model trained on one region to another region. To support mobility-related location adaptation experiments, we introduce the Dynamic Traffic Speeds++ (DTS++) dataset, an extension of DTS to include a new city, Cincinnati, OH. DTS++ is constructed in a similar manner to DTS and includes one year of historical traffic data collected from Uber Movement Speeds~\cite{movement}. This mobility data is paired with \num[group-separator={,}]{11137} overhead images ($1024 \times 1024$) at approximately $0.3$ meters per-pixel.

\setlength{\intextsep}{3pt}
\begin{wraptable}{r}{7.0cm}
  \centering
  \caption{An example of location adaptation using DTS++. When adapting our approach from New York City (NYC) to Cincinnati (Cincy), fine-tuning the geo-temporal positional encoding (GTPE) module dramatically improves performance.}

  \setlength\tabcolsep{2pt}
  
  \begin{tabular}{@{}cccSSS@{}}
    \toprule
    Train & Test & GTPE & \multicolumn{1}{c}{RMSE} & \multicolumn{1}{c}{MAE} & \multicolumn{1}{c}{$R^2$}  \\
    \bottomrule
    NYC & Cincy & {\em original}   & 24.06 & 18.84 & 0.04 \\
    NYC & Cincy & {\em fine-tuned} & 12.29 &  9.18 & 0.75 \\
    \bottomrule
  \end{tabular}
  
  \label{tbl:evaluation_la}
\end{wraptable}

Using DTS++, we conducted an initial experiment to evaluate how our approach, trained on New York City (NYC), performs when adapted to Cincinnati (Cincy). We show the results of this experiment in \tabref{evaluation_la}. As expected, performance deteriorates as Cincy has vastly different spatiotemporal mobility patterns than NYC. For example, the average speed across all road segments in DTS++ for NYC is 18.93 km/h while Cincy is 31.51 km/h. In addition, we show that fine-tuning our geo-temporal positional encoding module (GTPE) on Cincy dramatically improves performance. This experiment simultaneously highlights the impact of GTPE on traffic speed estimation and shows that fine-tuning only the associated $\sim 65$k parameters can be an efficient way of adapting models to new locations. 

In summary, while location adaptation is not the primary focus of this work, it is a nascent and important research direction and our hope is that DTS++ helps enable future studies related to mobility-related location adaptation.

\section{Applications}

\paragraph{Creating Dense City-Scale Traffic Models} Though empirical traffic data has become increasingly available, not all roads are traversed at all times. This presents challenges for downstream applications which rely on data from transport modeling, as they are limited to regions/times where data is available, or must collect their own data. Our approach presents an alternative as it can be used to create dense city-scale traffic models. \figref{cartoon} (left) shows available historical traffic data from the Dynamic Traffic Speeds dataset (Brooklyn, Monday 8am). As observed, many roads are missing empirical speed data (white) as they were not traversed at the given time. \figref{cartoon} (right) shows how our approach can be used to generate a dense model of traffic flow due to its ability to generalize across space and time.

\setlength{\intextsep}{\defaultintextsep}
\begin{figure}[t]
    \centering
    \newcommand{\centered}[1]{\begin{tabular}{l} #1 \end{tabular}}
    \setlength\tabcolsep{1pt}
    \begin{tabular}{cccc}
        \centered{\includegraphics[width=.19\linewidth]{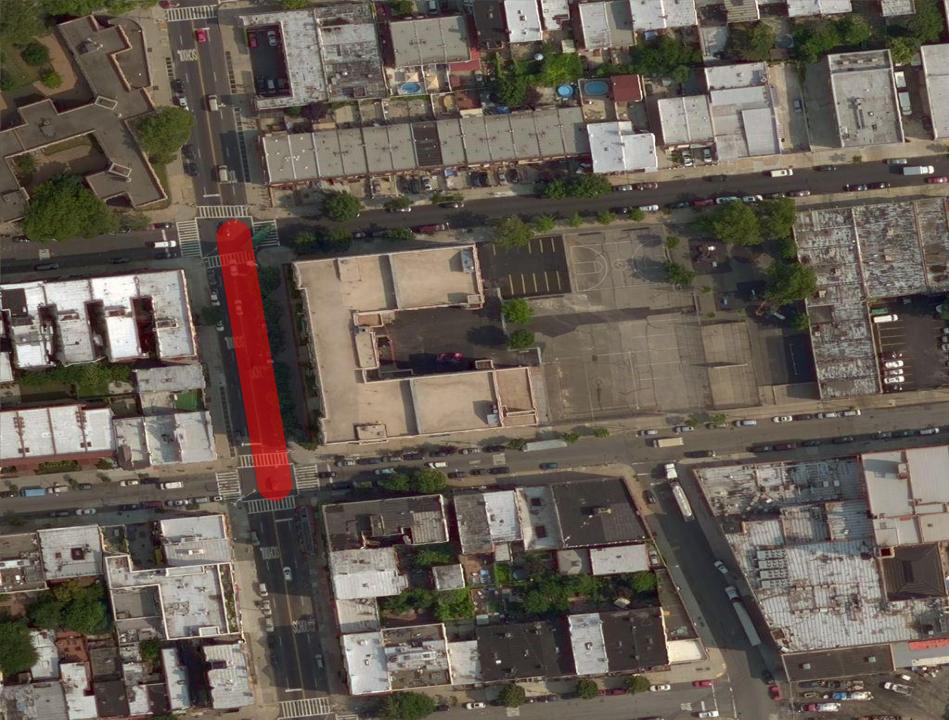}} &
        \centered{\includegraphics[width=.29\linewidth]{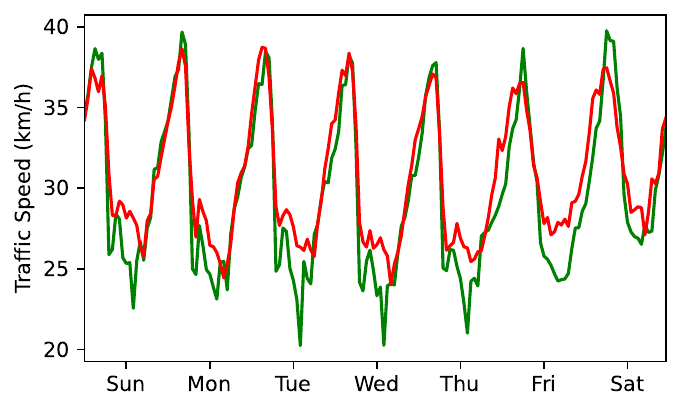}} &
        \centered{\includegraphics[width=.19\linewidth]{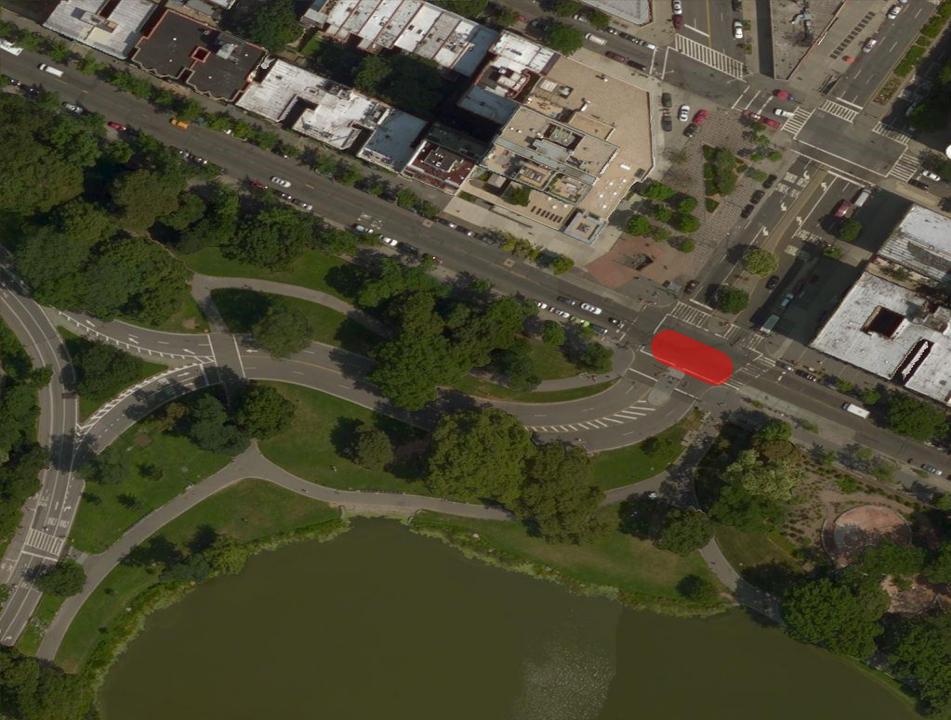}} &
        \centered{\includegraphics[width=.29\linewidth]{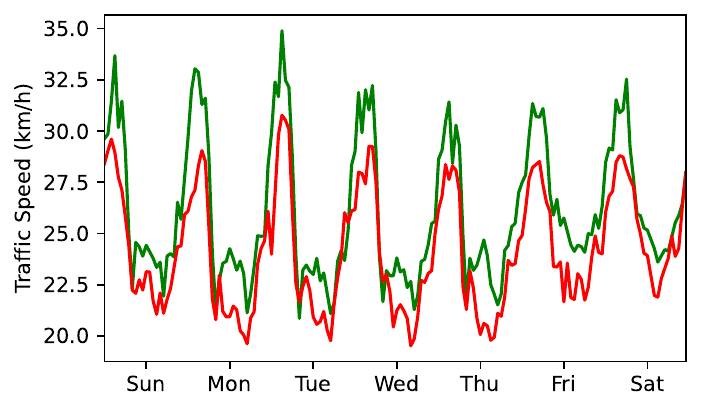}} \\
        
        \centered{\includegraphics[width=.19\linewidth]{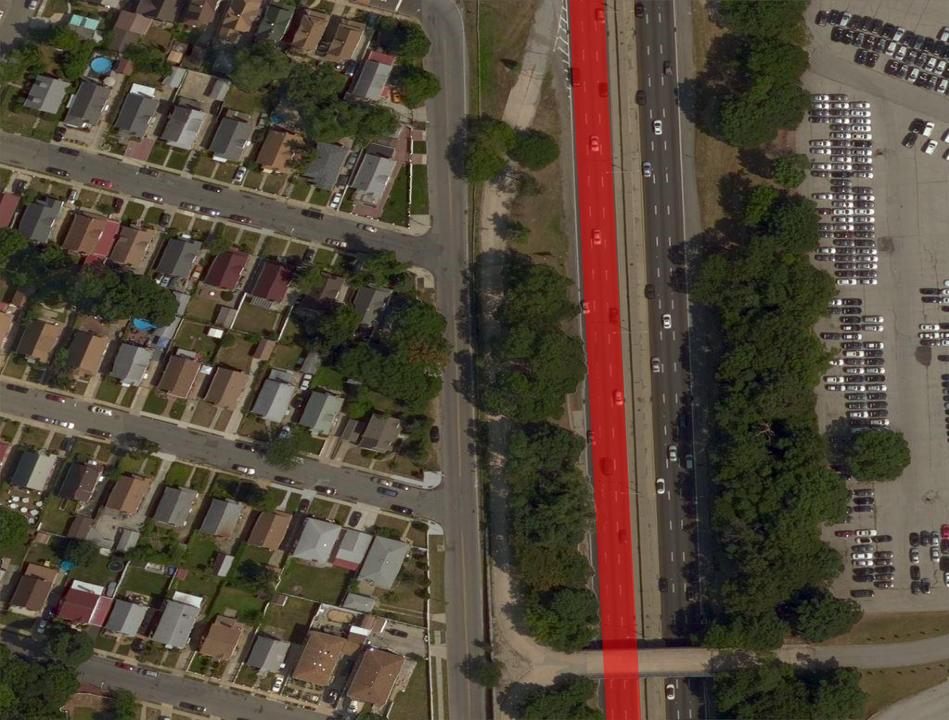}} &
        \centered{\includegraphics[width=.29\linewidth]{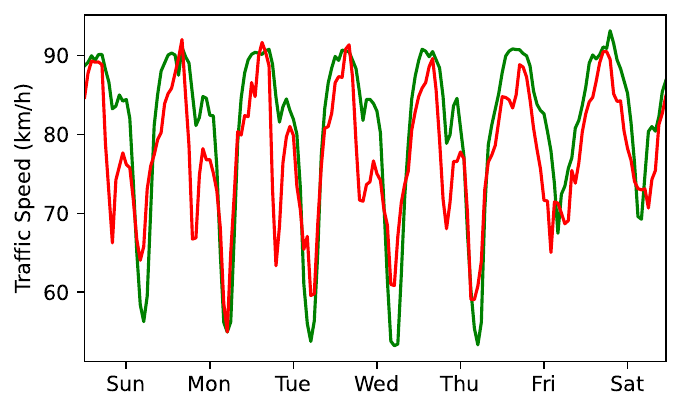}} &
        \centered{\includegraphics[width=.19\linewidth]{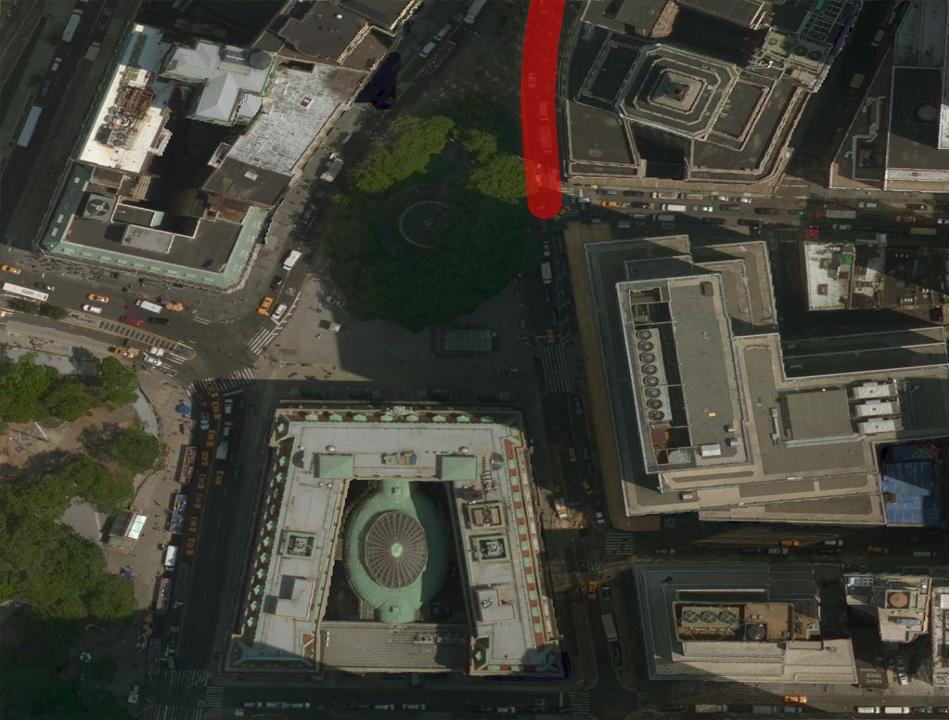}} &
        \centered{\includegraphics[width=.29\linewidth]{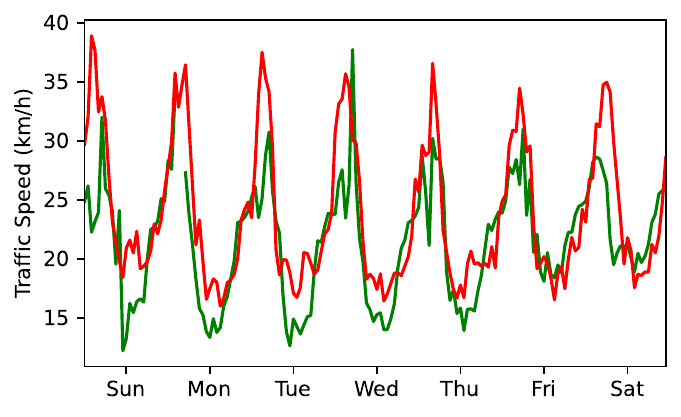}} \\
    \end{tabular}
    \caption{Visualizing traffic speed predictions for individual road segments versus time. (left) A road segment shown in red. (right) Predictions from our approach (red) versus historical traffic data (green).}
    \label{fig:segments}
\end{figure}

\paragraph{Modeling Traffic Flow} Our approach can be used to model spatiotemporal traffic patterns for individual road segments. To demonstrate this, we analyze how predictions from our approach on unseen road segments compare to historical traffic data. \figref{segments} shows the results of this experiment. The x-axis represents time, with each day representing a 24 hour interval. Our approach (red) is able to capture temporal trends in historical traffic data (green).

\begin{figure}[t]
    \centering
    \includegraphics[width=.49\linewidth]{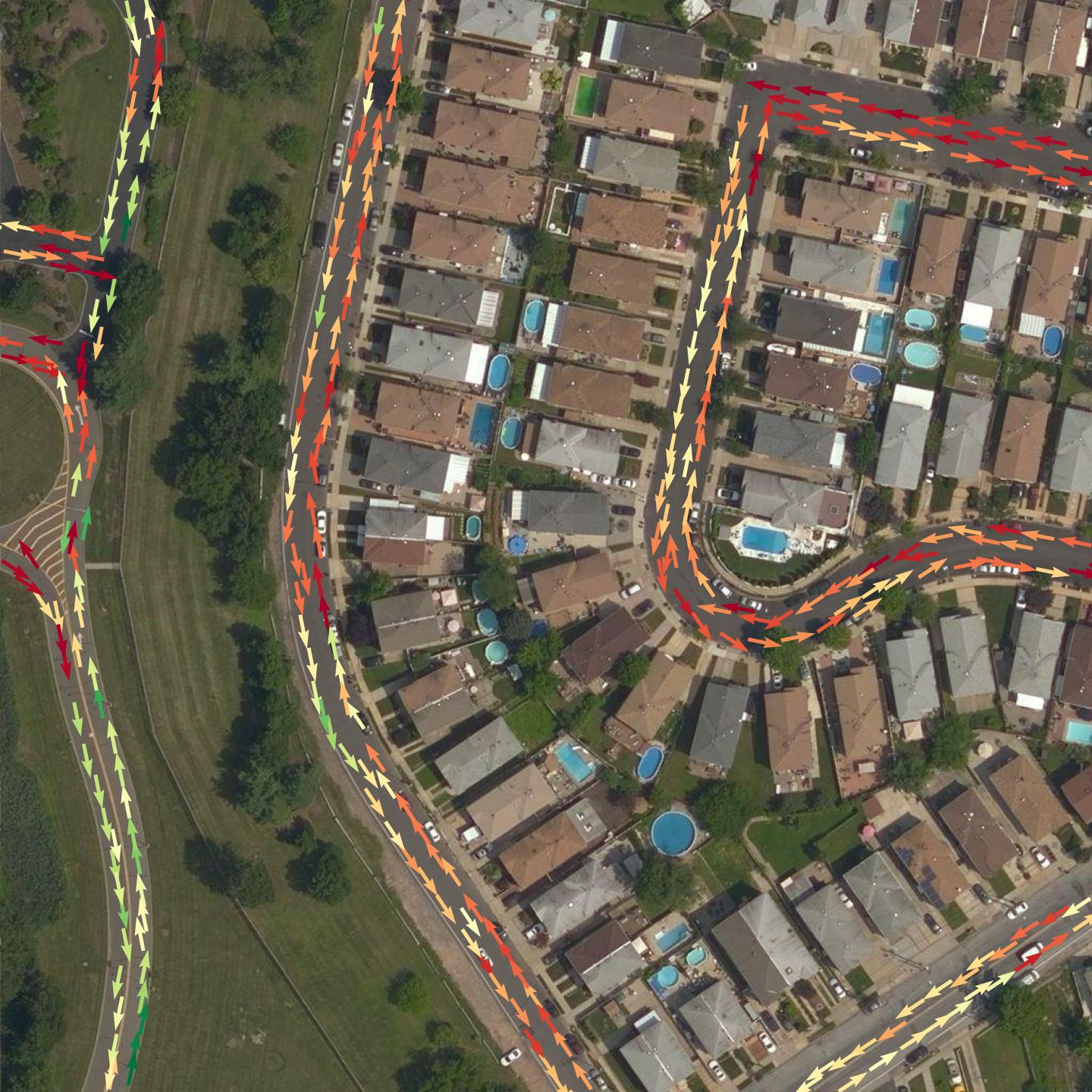}
    \includegraphics[width=.49\linewidth]{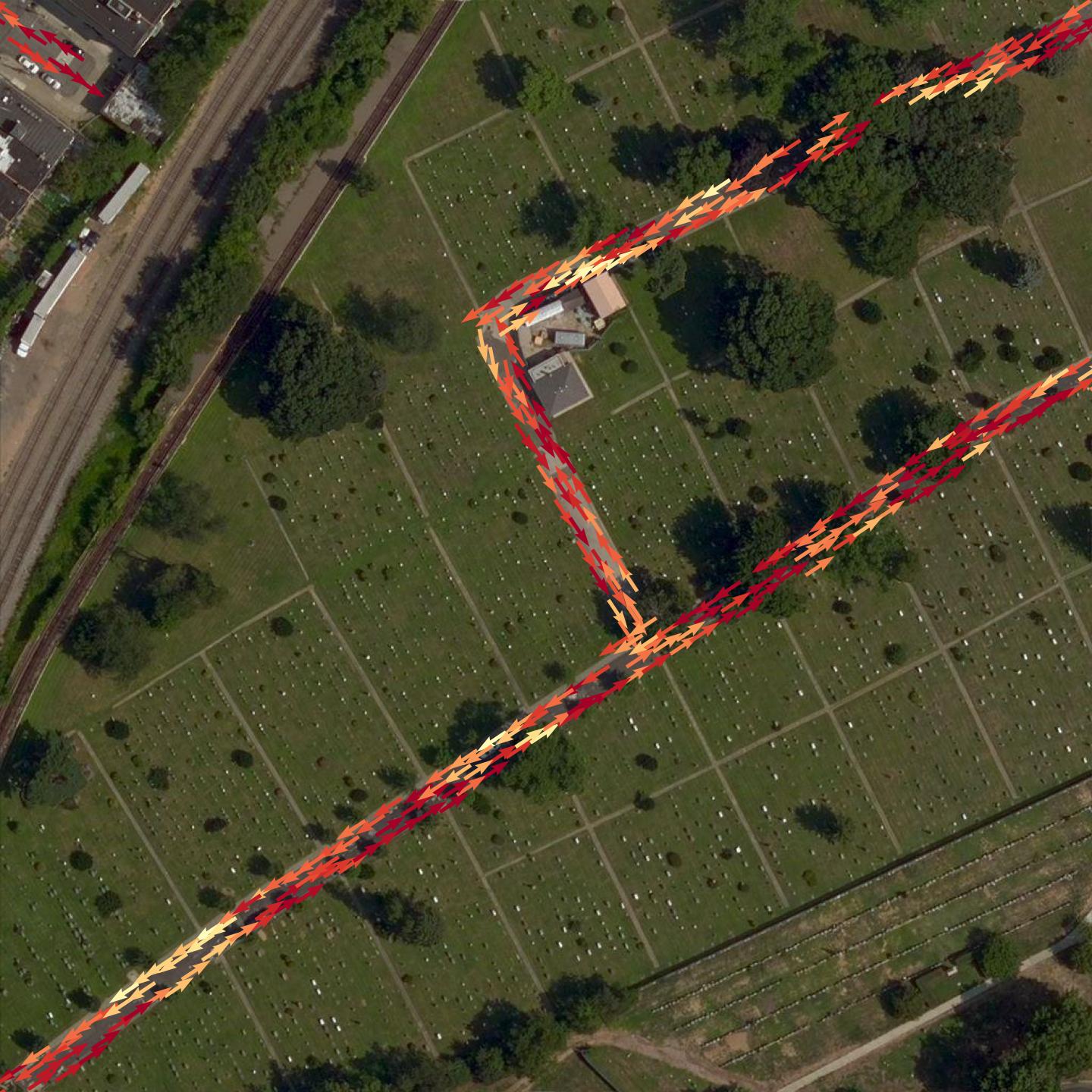}
    \caption{The output of our approach on unseen locations visualized as a local motion model. We sample from locations predicted as road and visualize corresponding orientation estimates as vectors, colored by the predicted speed. Green (red) corresponds to fast (slow) traffic speeds.}
    \label{fig:generalizing}
\end{figure}

\paragraph{Generalizing to Novel Locations} Our approach can be thought of as estimating a local motion model from an overhead image which describes how the environment is traversed. \figref{generalizing} visualizes how our method can be applied to novel locations to generate a local motion model. This allows our approach to be applied to locations where transport data is incomplete or unavailable, a common occurrence. For example, road networks are not always accurate for a given location (e.g., unmapped or changed). Further, traffic speed data is sparse and not always available for all roads. The generalizability of our approach is made possible by the inclusion of the auxiliary tasks of road segmentation and orientation estimation, in addition to their performance benefit from multi-task learning. 

\section{Conclusion}

Our goal was to understand city-scale mobility patterns using overhead imagery, a task we refer to as image-driven traffic modeling. We proposed a multi-modal, multi-task transformer-based segmentation architecture and showed how it can be used to create dense city-scale traffic models. Our method has several key components, including a probabilistic formulation for naturally modeling variations in traffic speeds, and a geo-temporal positional encoding module for integrating geo-temporal context. Extensive experiments using the Dynamic Traffic Speeds (DTS) dataset demonstrate how our approach significantly improves the state-of-art in traffic speed estimation. Finally, we introduced the DTS++ dataset to support motion-related location adaptation studies across diverse cities. Our hope is that these results continue to demonstrate the real-world utility of image-driven traffic modeling.

\bibliographystyle{splncs04}
\bibliography{biblio}

\null
\vspace*{-27pt}
\begin{center}
  \textbf{\Large Supplemental Material : \\ Probabilistic Image-Driven Traffic \\ Modeling via Remote Sensing}
\end{center}
\vspace*{24pt}
\thispagestyle{empty}
\setcounter{section}{0}
\setcounter{equation}{0}
\setcounter{figure}{0}
\setcounter{table}{0}
\makeatletter
\renewcommand{\theequation}{S\arabic{equation}}
\renewcommand{\thefigure}{S\arabic{figure}}
\renewcommand{\thetable}{S\arabic{table}}

This document contains additional details and experiments related to our methods.

\section{Dynamic Traffic Speeds++ (DTS++)}

\begin{figure}[!b]
    \centering
    \frame{\includegraphics[width=.8\linewidth]{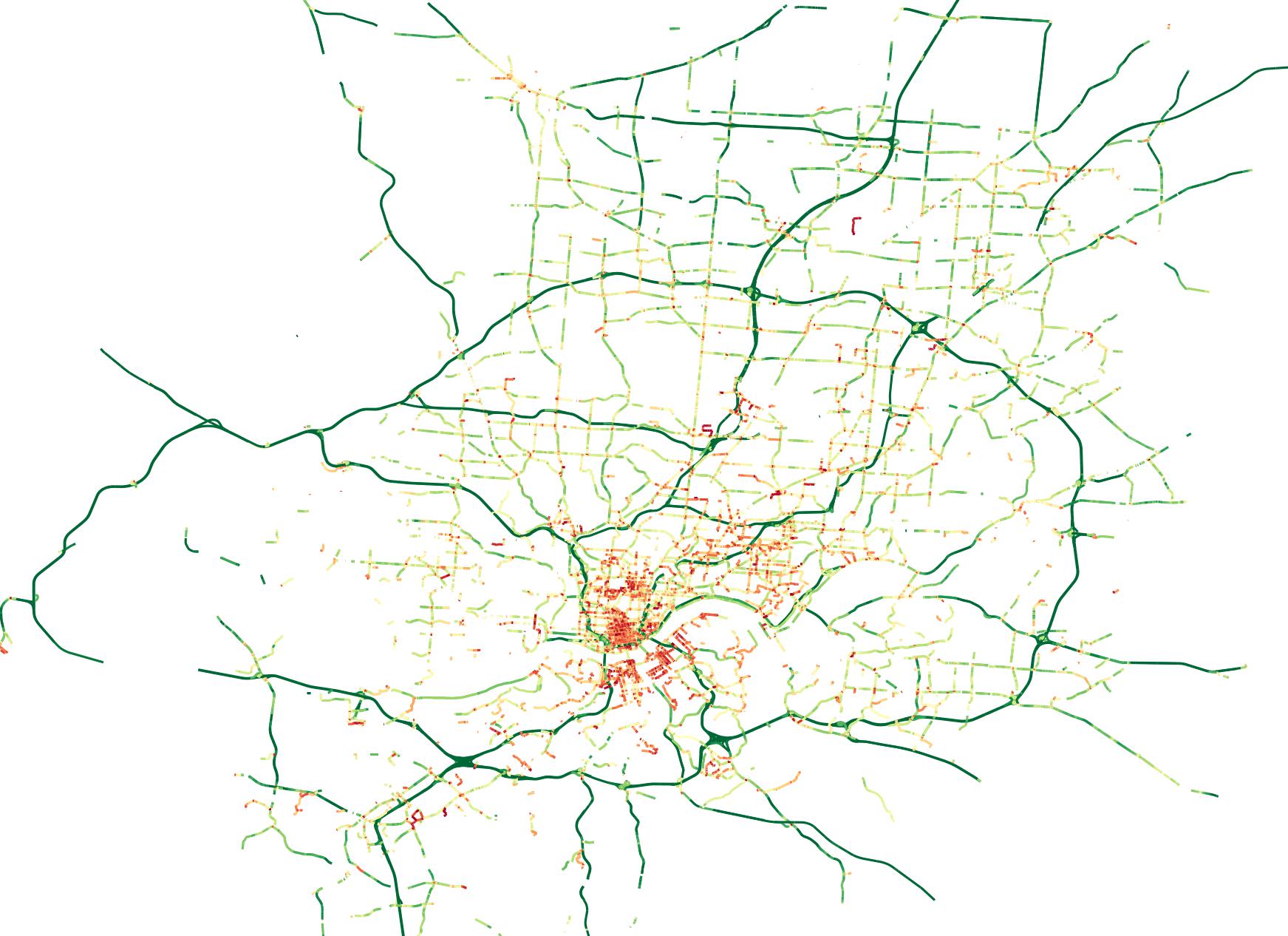}}
    \caption{Example traffic speed data from DTS++ for Cincinnati (road segment speeds averaged over time). Green (red) corresponds to fast (slow) traffic speeds.}
    \label{fig:s_cincinnati}
\end{figure}

We introduce DTS++, an extension of the Dynamic Traffic Speeds (DTS) dataset~\cite{workman2020dynamic} to include an additional city, Cincinnati, OH. To construct DTS++, we emulated the process used to create DTS, which we outline here. First, we obtained a year of historical traffic speed data (2018) from Uber Movement Speeds~\cite{movement}, a dataset of publicly available aggregated speed data derived from Uber rideshare trips. The speed data is provided as averages over road segments at an hourly resolution, with the underlying road network described using OpenStreetMap. \figref{s_cincinnati} gives a visual overview of Cincinnati's spatial coverage (traffic speeds averaged across time).

To align the historical traffic speed data with overhead imagery, we first generated a set of non-overlapping spherical Mercator tiles (XYZ style) that intersect a bounding box around Cincinnati. We identified tiles containing road segments with available speed data, and for each, downloaded an overhead image from Bing Maps (at a resolution of approximately $0.3$ meters per-pixel). This process resulted in \num[group-separator={,}]{11137} overhead images ($1024 \times 1024$) with associated road geometries and traffic speed data. To support model training and evaluation, these non-overlapping tiles are randomly split into 85\% training, 5\% validation, and 10\% testing, such that road segments observed during testing are not observed during training. 

\begin{figure}[t]
    \centering
    \newcommand{\centered}[1]{\begin{tabular}{l} #1 \end{tabular}}
    \setlength\tabcolsep{1pt}
    \begin{tabular}{cc}
        \centered{\includegraphics[width=.439\linewidth]{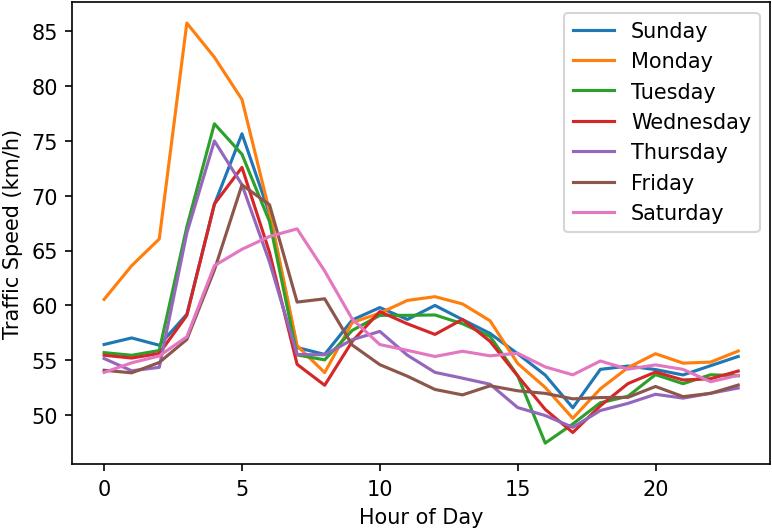}} &
        \centered{\includegraphics[width=.54\linewidth]{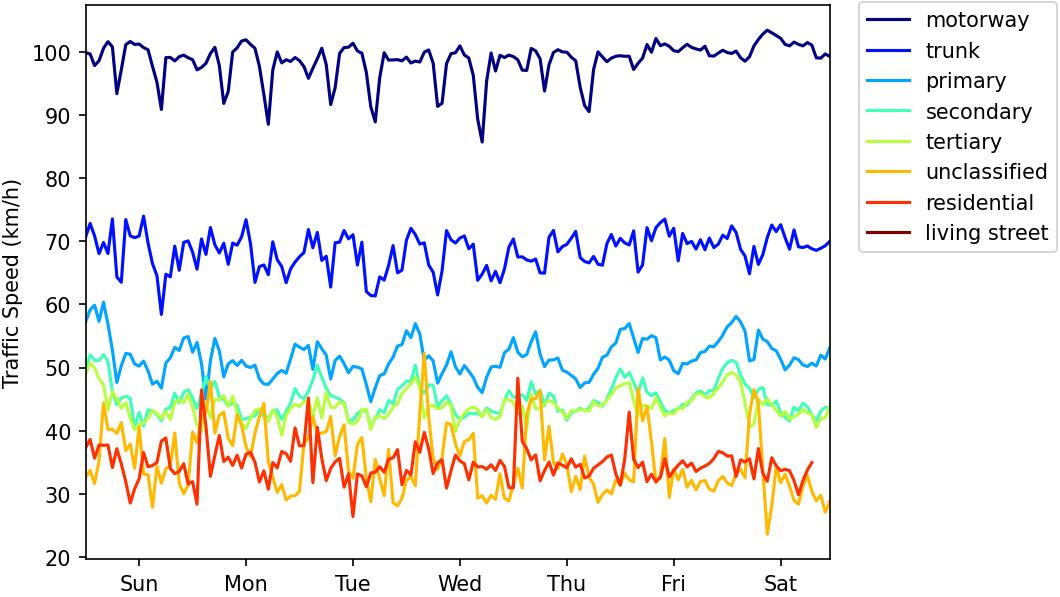}} \\
    \end{tabular}
    \caption{(left) Visualizing the average road segment speed over time in Cincinnati for different days. (right) Visualizing the average road segment speed against time in Cincinnati, where each road is categorized by its OpenStreetMap road type classification.}
    \label{fig:s_speed_type_vs_time}
\end{figure}

In \figref{s_speed_type_vs_time} (left), we visualize the average road segment speed versus time for individual days of the week. Additionally, \figref{s_speed_type_vs_time} (right) shows the average road segment speed for different types of roads using OpenStreetMap's road type classification. DTS++ extends the DTS dataset by adding a new city, Cincinnati, essentially doubling the size of the dataset and providing support for location adaptation experiments. Compared to New York City (from DTS), Cincinnati has very different appearance characteristics as well as spatiotemporal mobility patterns. For example, the maximum observed traffic speed in Cincinnati for a motorway is $\sim100$ km/h, where in New York City it is $\sim80$ km/h. 

\section{Fine-Grained Visual Categorization}

\begin{figure}[t]
    \centering

    \begin{tblr}{colspec={X[1,c,m] *{3}{X[2,c,m]}},width=1\linewidth,rowsep=1pt,colsep=5pt,vspan=even,}

        & January & May & September \\

        \includegraphics[valign=m,width=1\linewidth]{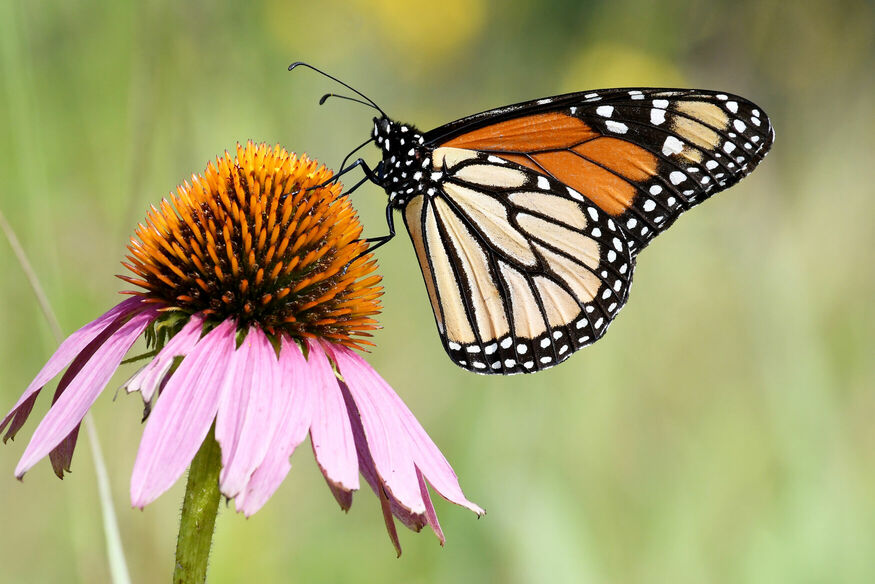} &
        \SetCell[r=2]{h}{\includegraphics[valign=m,width=1\linewidth]{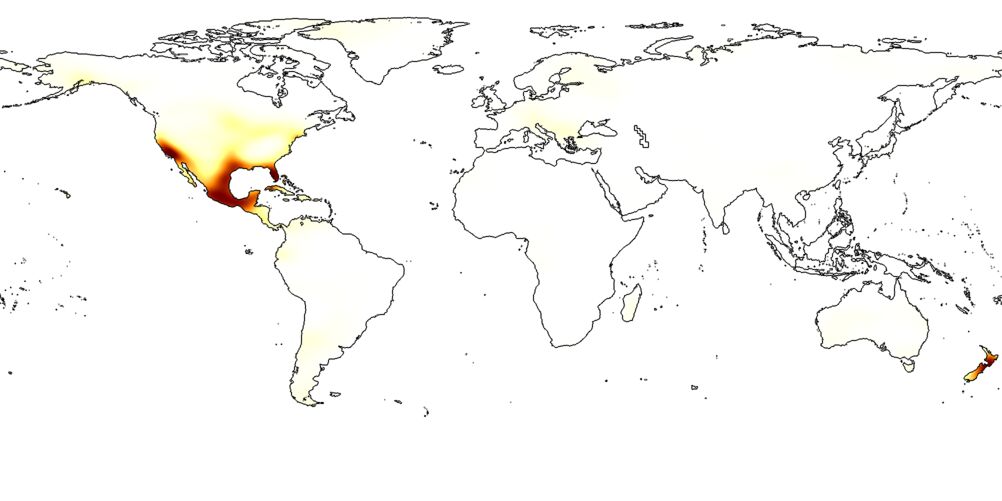}} &
        \SetCell[r=2]{h}{\includegraphics[valign=m,width=1\linewidth]{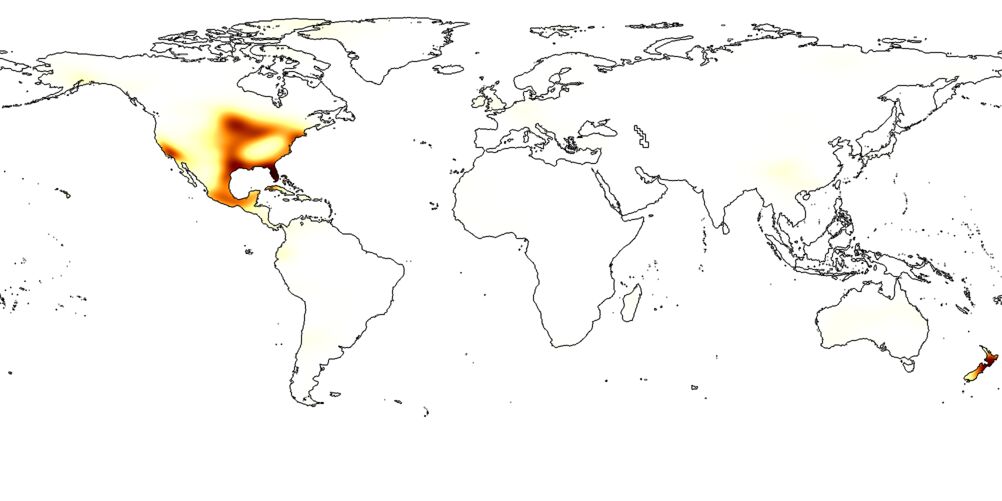}} &
        \SetCell[r=2]{h}{\includegraphics[valign=m,width=1\linewidth]{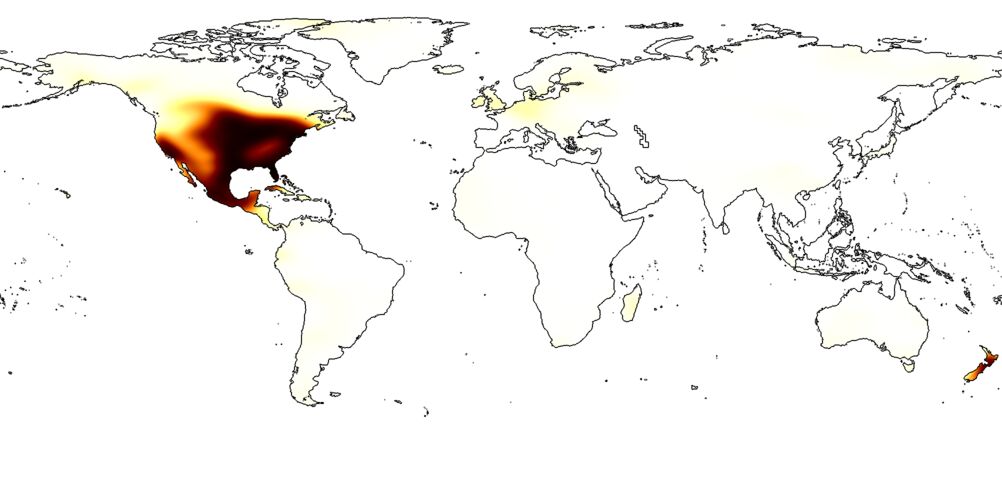}} \\
        {\small Monarch Butterfly} & & & \\

        \includegraphics[valign=m,width=1\linewidth]{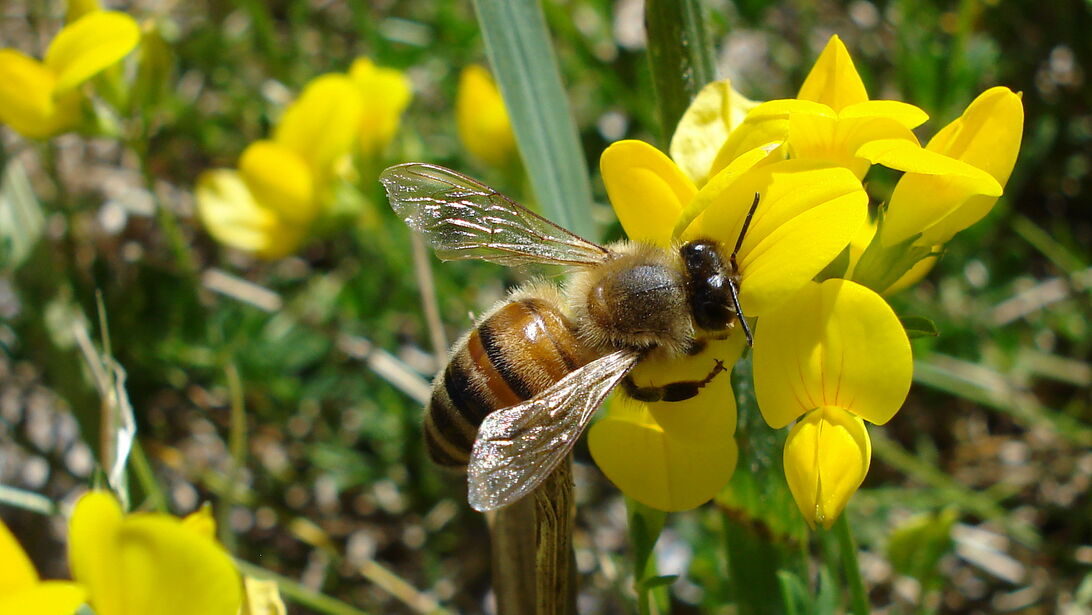} &
        \SetCell[r=2]{h}{\includegraphics[valign=m,width=1\linewidth]{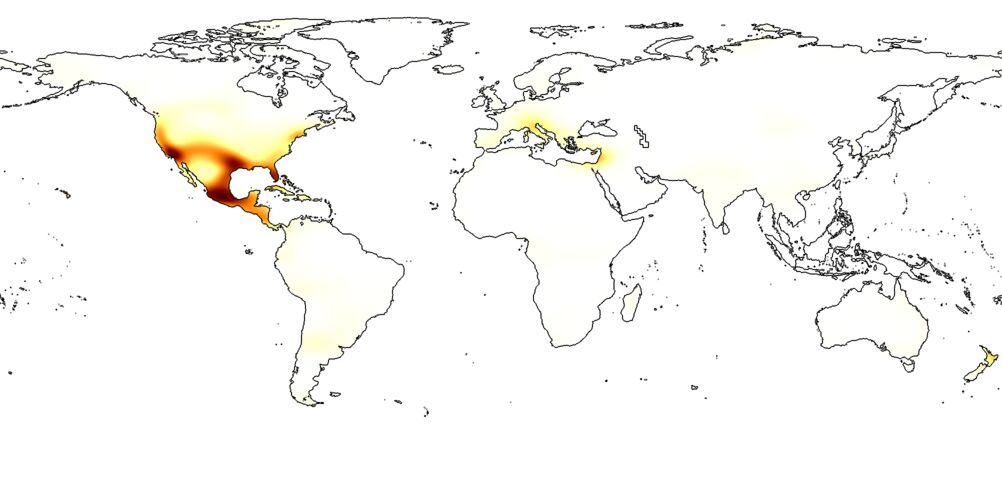}} &
        \SetCell[r=2]{h}{\includegraphics[valign=m,width=1\linewidth]{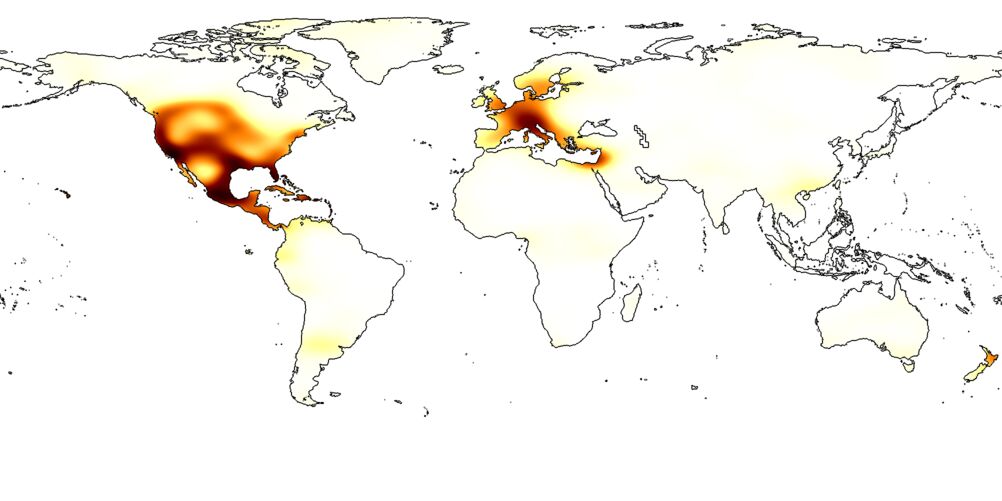}} &
        \SetCell[r=2]{h}{\includegraphics[valign=m,width=1\linewidth]{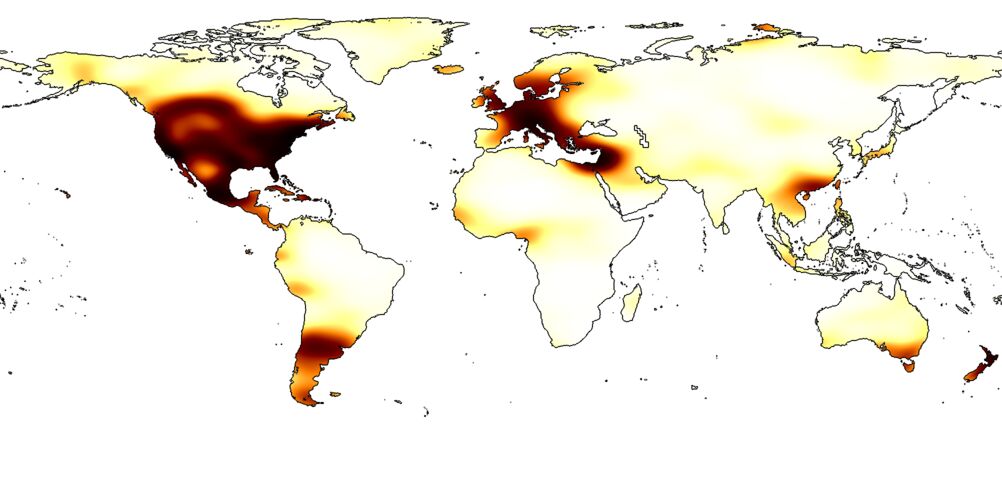}} \\
        {\small Western Honey Bee} & & & \\
        
    \end{tblr}

    \caption{Visualizing the spatiotemporal prior, learned using the GTPE module, for several species in the iNaturalist 2018 dataset. Darker colors correspond to higher likelihood of the species appearing at that location.}
    \label{fig:s_qualitative_presence}
\end{figure}

We investigate the performance of our geo-temporal positional encoding (GTPE) module on a different task, fine-grained visual categorization, using the iNaturalist 2018 dataset~\cite{van2018inaturalist}. This dataset contains over \num[group-separator={,}]{450000} images of birds (approximately \num[group-separator={,}]{8000} unique species) collected from the iNaturalist social network. The task is to perform large-scale, fine-grained species classification in the scenario where there are many visually similar species and high class imbalance.

For this experiment, we take advantage of the framework introduced by Aodha et al.~\cite{mac2019presence}. The authors proposed to learn a spatiotemporal prior, $P(y|\boldsymbol{x})$, conditioned on location and time context, $\boldsymbol{x}$, that estimates the probability that a given object category, $y$, occurs at that location. The spatiotemporal prior is combined with the output of an image classifier, $P(y|I)$. The results show that including the spatiotemporal prior helps disambiguate fine-grained categories. We take advantage of this framework, but use our proposed geo-temporal positional encoding module to learn the spatiotemporal prior, instead of their approach which uses a MLP consisting of several residual layers and a final output embedding layer (each of 256 dimensions). 

We start from public code made available by the authors. The only changes we make to the base training configuration are to set the learning rate to $10^{-4}$ and remove the learning rate decay policy (for fairness across methods). When integrating GTPE, we use the same parameterization for location/time as the authors and set the hidden/output dims to 256 to follow precedent. Note that our goal here is not to achieve state-of-the-art performance on this benchmark, but rather to understand the impact of our GTPE module against a seminal approach for encoding location/time context. 

\tabref{s_ablation_gtpe} shows the result of this experiment in terms of Top-1, Top-3 and Top-5 accuracy. For all methods, we keep training parameters consistent, only changing which context (location, {\em loc}, or time, {\em time}) is made available. For GTPE, {\em loc}, {\em time}, and {\em loc+time}, correspond to the enabling or disabling of individual pathways inside the module. GTPE outperforms the baseline, including both a variant trained locally with the same training configuration, and the original results presented in the paper. \figref{s_qualitative_presence} shows an example of the spatiotemporal distributions learned by GTPE for several species.

\begin{table}
  \small
  \centering
  \caption{Ablation study of our geo-temporal positional encoding on iNaturalist 2018.}

  \setlength\tabcolsep{6pt}
  \begin{tabular}{@{}llccc@{}}
    \toprule
    Method & Context & Top-1 & Top-3 & Top-5 \\
    \bottomrule
    \multirow{2}{*}{\cite{mac2019presence} (paper)}
    & {\em loc}       & 72.41 & 87.19 & 90.60 \\
    & {\em loc, time} & 72.68 & 87.26 & 90.79 \\
    \midrule
    \multirow{3}{*}{\cite{mac2019presence} (local)} 
    & {\em loc}       & 71.62 & 86.62 & 90.27 \\
    & {\em loc, time} & 71.53 & 86.75 & 90.28 \\
    \midrule
    \multirow{3}{*}{Ours}
    & {\em loc}                   & 72.64 & 87.21 & \textbf{90.80} \\
    & {\em loc, time}             & 72.66 & 87.28 & 90.70 \\
    & {\em loc, time, loc+time} & \textbf{72.73} & \textbf{87.31} & 90.75 \\
    \bottomrule
  \end{tabular}
  
  \label{tbl:s_ablation_gtpe}
\end{table}

\section{Additional Experiments}

\subsection{Extended Ablation}

To evaluate our choice of backbone architecture, we trained a variant of our approach replacing our encoder (which combines convolutional layers and attention layers) with the hierarchical transformer encoder from SegFormer~\cite{xie2021segformer}, omitting contextual inputs (location and time). This is equivalent to our image-only variant. When evaluated on New York City, the performance of this baseline is 10.51 RMSE, 7.79 MAE, and 0.48 $R^2$, which is worse than our image-only variant (10.28, 7.70, and 0.50 respectively), but inline with expectations. Though these results show our choice of backbone architecture to be superior for image-driven traffic modeling, we would note that our proposed components (e.g., GTPE, probabilistic formulation) could conceivably be combined with any backbone architecture. 

\subsection{Visualizing Geo-Temporal Context}

\begin{figure}[b]
    \centering
    \includegraphics[width=.8\linewidth]{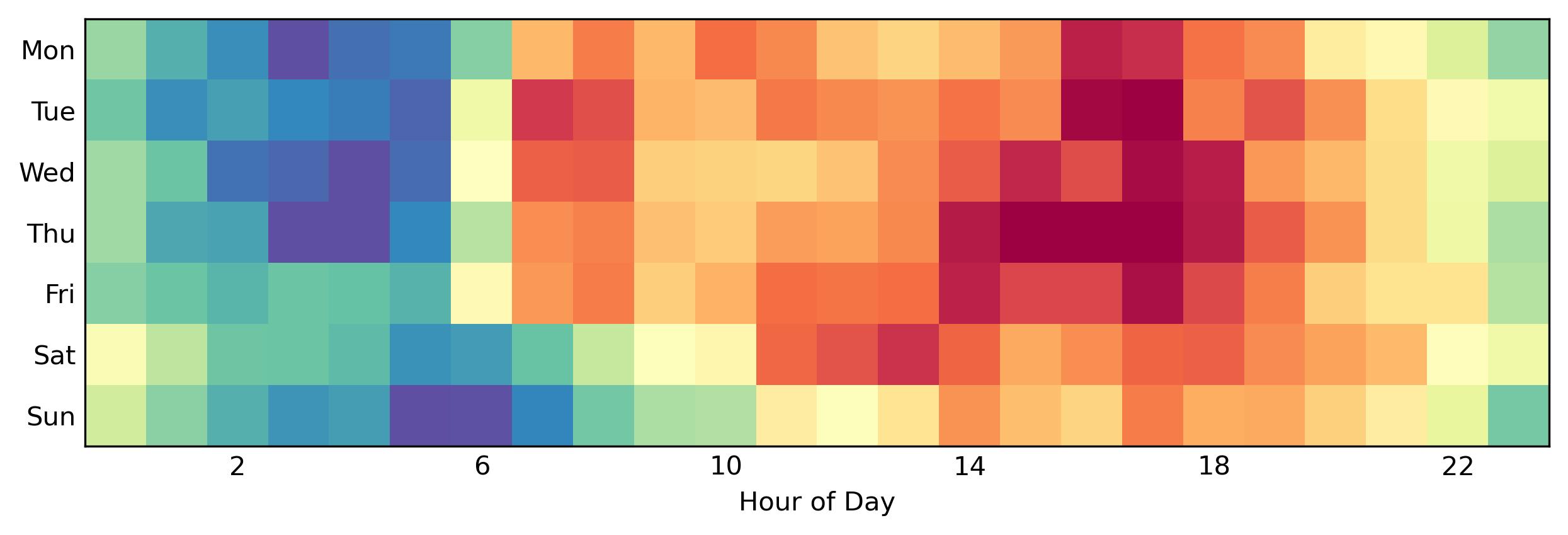}
    \caption{Visualizing the time embedding learned by GTPE.}
    \label{fig:s_time}
\end{figure}

Our method integrates location and time through distinct pathways in the GTPE module which can be thought of as 64-dimensional embeddings. To visualize the learned feature representations for time, we construct a false color image using principal component analysis (PCA), shown in \figref{s_time}. For every time combination (day of week, hour of day), we extract its feature representation from the time embedding, and stack them. This results in a matrix of size $168 \times 64$. We apply PCA to reduce the feature dimensionality and reshape to $7 \times 24$ (days by hours). The resulting image represents time-dependent traffic patterns learned by our approach. 

As observed, red correlates with higher traffic activity and blue depicts lower. For example, there is less activity in the evenings (midnight-7am), with an increase in activity at 8am across all days as people tend to become active at that time. The visualization also shows that the time embedding captures additional trends, such as: 1) peak traffic around 5pm-6pm across all days, 2) Sunday having reduced traffic throughout the entire day, and 3) increased activity early on Saturday and Sunday mornings. Overall, this visualization shows that our proposed GTPE module is learning to relate geo-temporal context to mobility patterns.

\subsection{Capturing Uncertainty in Traffic Speeds}

\begin{figure}
    \centering
    \newcommand{\centered}[1]{\begin{tabular}{l} #1 \end{tabular}}
    \setlength\tabcolsep{1pt}
    \begin{tabular}{cccc}
        \centered{\includegraphics[width=.196\linewidth]{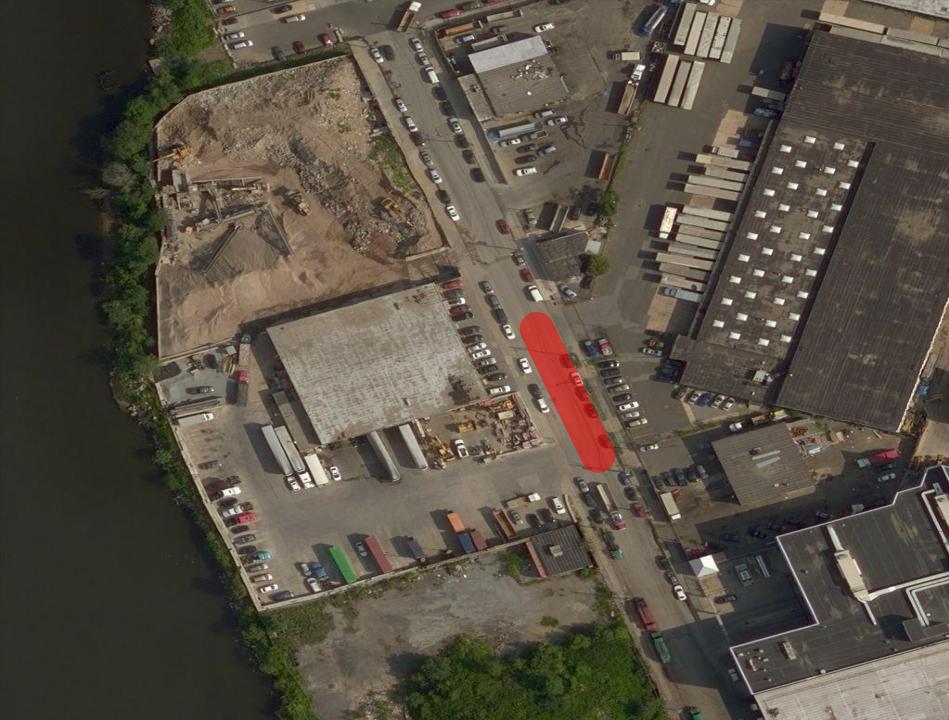}} &
        \centered{\includegraphics[width=.2842\linewidth]{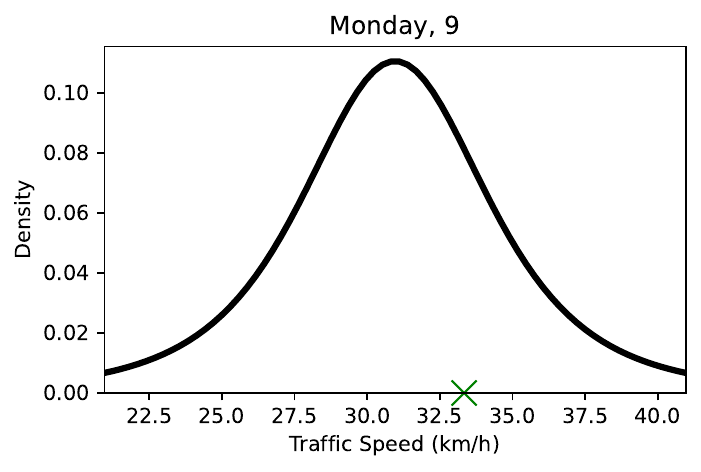}} &
        \centered{\includegraphics[width=.196\linewidth]{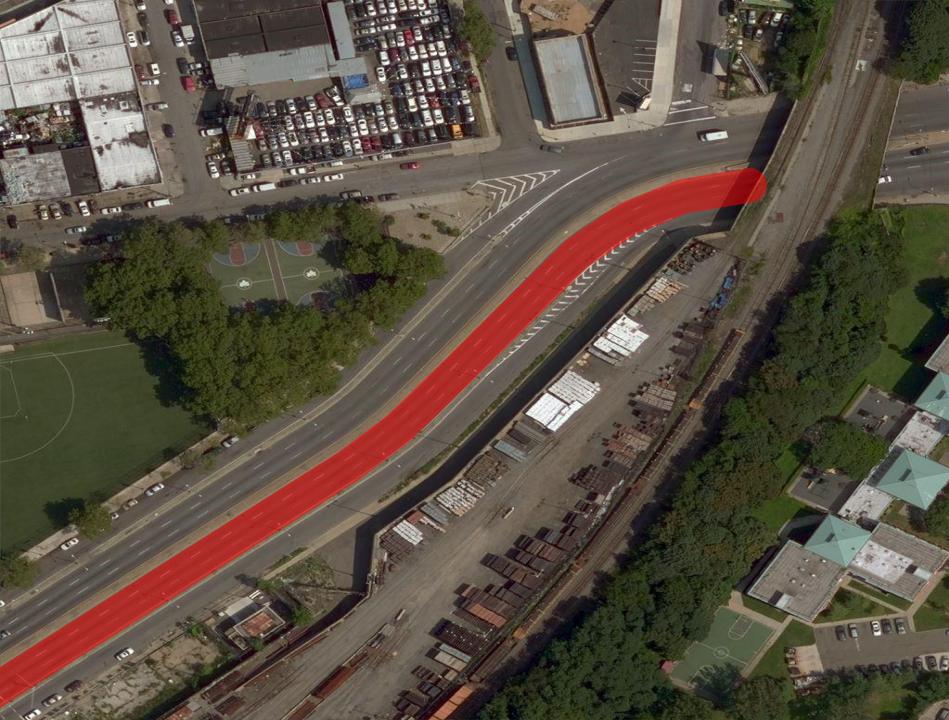}} &
        \centered{\includegraphics[width=.2842\linewidth]{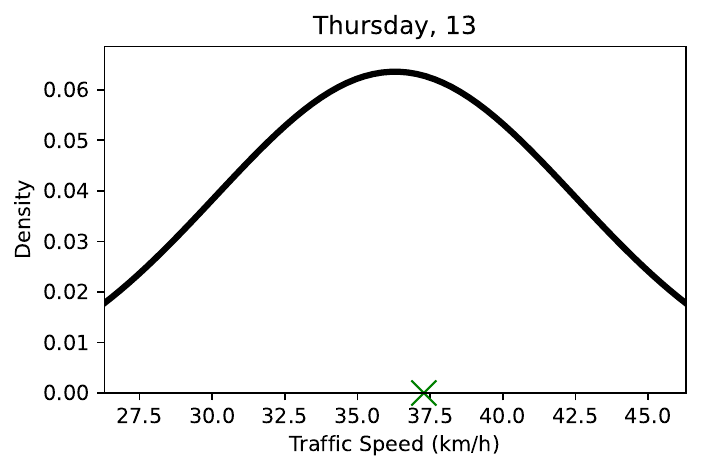}} \\
        
        \centered{\includegraphics[width=.196\linewidth]{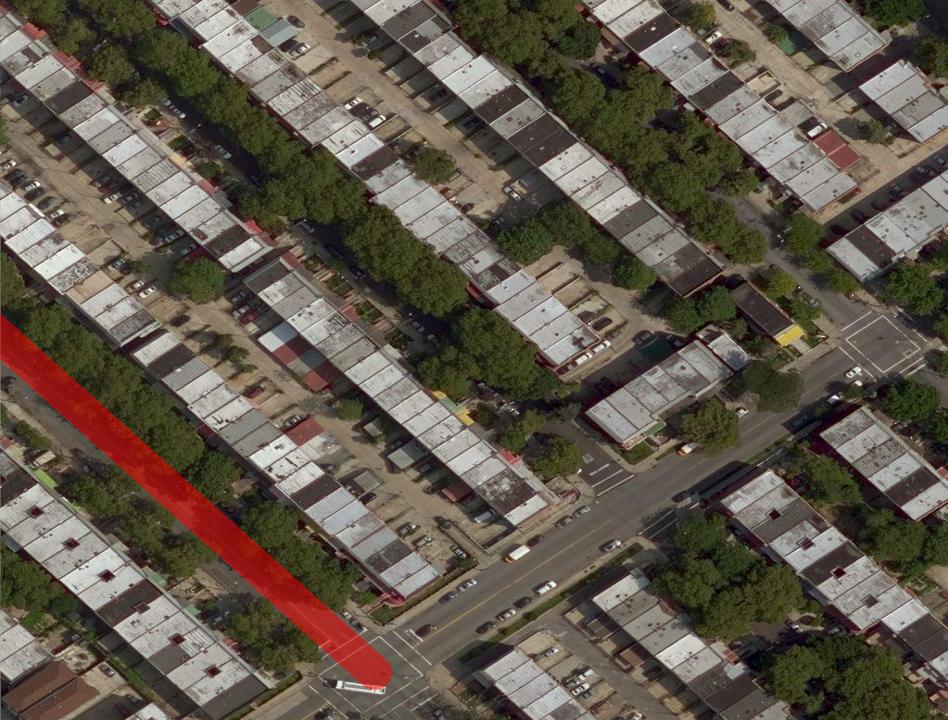}} &
        \centered{\includegraphics[width=.2842\linewidth]{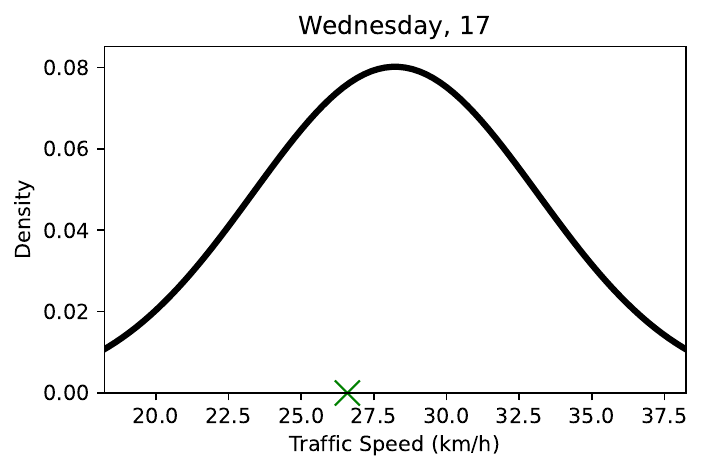}} &
        \centered{\includegraphics[width=.196\linewidth]{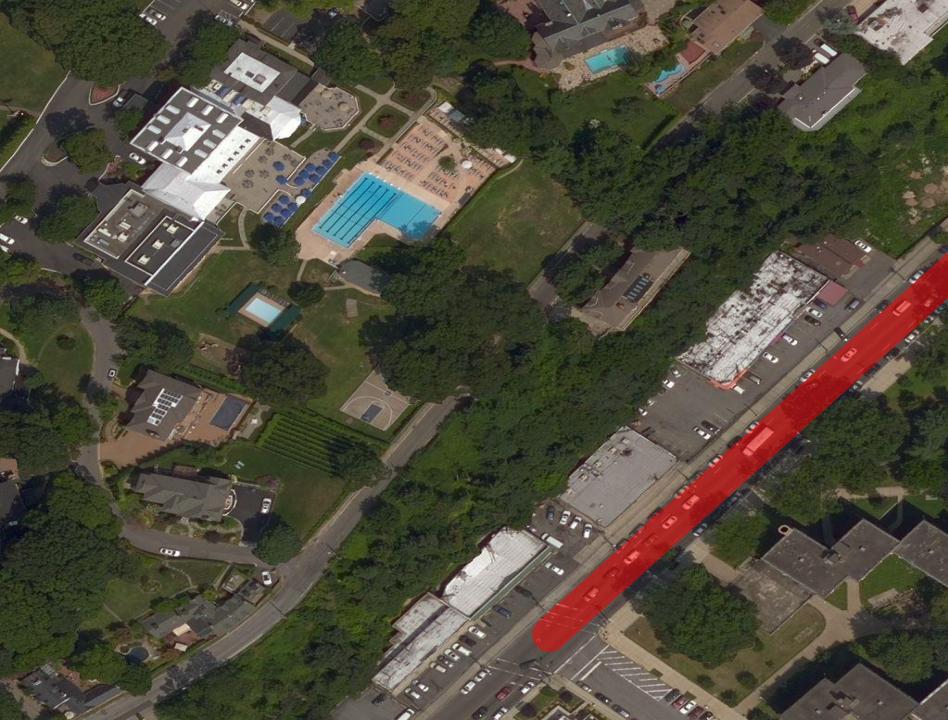}} &
        \centered{\includegraphics[width=.2842\linewidth]{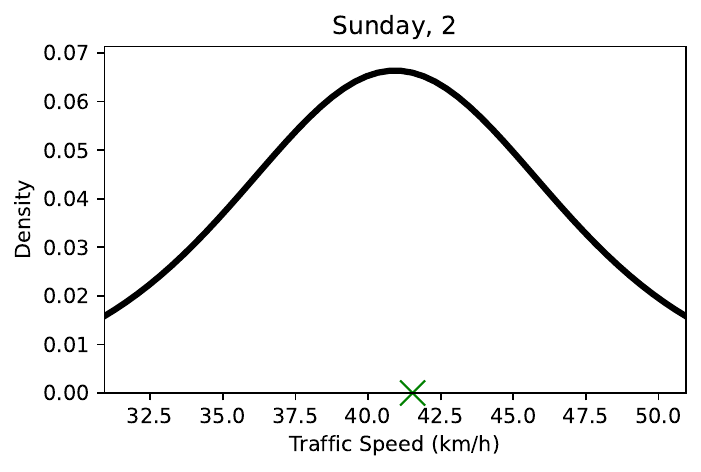}} \\
    \end{tabular}
    \caption{Our method outputs prior distributions over traffic speeds that implicitly capture uncertainty. (right) Output from our approach visualized as a probability density function corresponds to the (left) road segment depicted in the image. The green $x$ represents the ground-truth traffic speed at the given time.}
    \label{fig:s_uncertainty}
\end{figure}

Our method implicitly models uncertainty in traffic speeds at a location/time. This is due to our probabilistic formulation, where instead of regressing traffic speeds directly, we estimate prior distributions over traffic speeds. The output of our approach are the location and scale parameters of a (per-pixel) Student's t-distribution, which are aggregated across an individual road segment and combined with the observed count of traffic observations to form a per-road-segment Student's t-distribution. During model training, we treat the ground-truth traffic speed for a given segment as a sample from the estimated distribution, and minimize negative log-likelihood. \figref{s_uncertainty} visualizes how our approach captures the underlying uncertainty in traffic speed for a given road segment.

\begin{figure}
    \centering

    \setlength\tabcolsep{1pt}
    
    \begin{tabular}{ccccc}        
        Image & Road (Label) & Road (Pred.) & Orient. (Label) & Orient. (Pred.) \\
        
        \includegraphics[width=.194\linewidth]{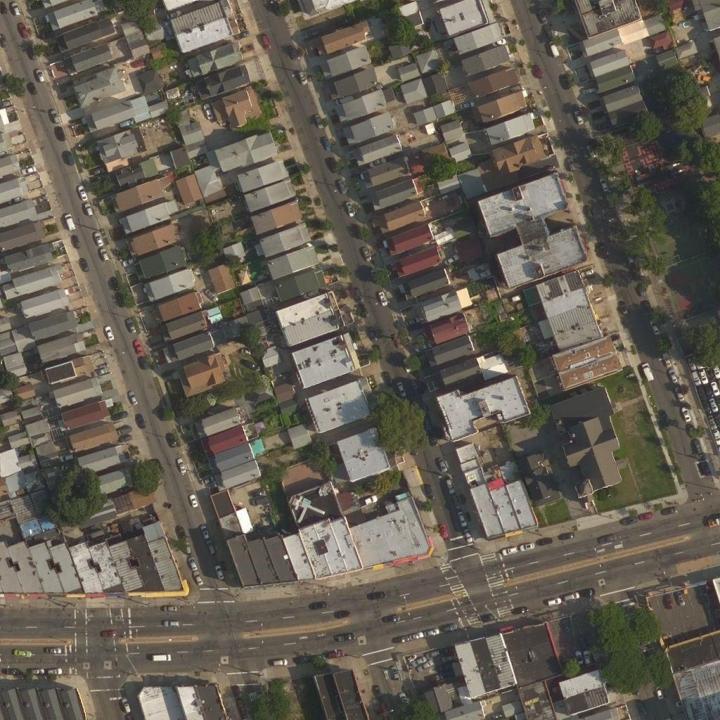} &
        \includegraphics[width=.194\linewidth]{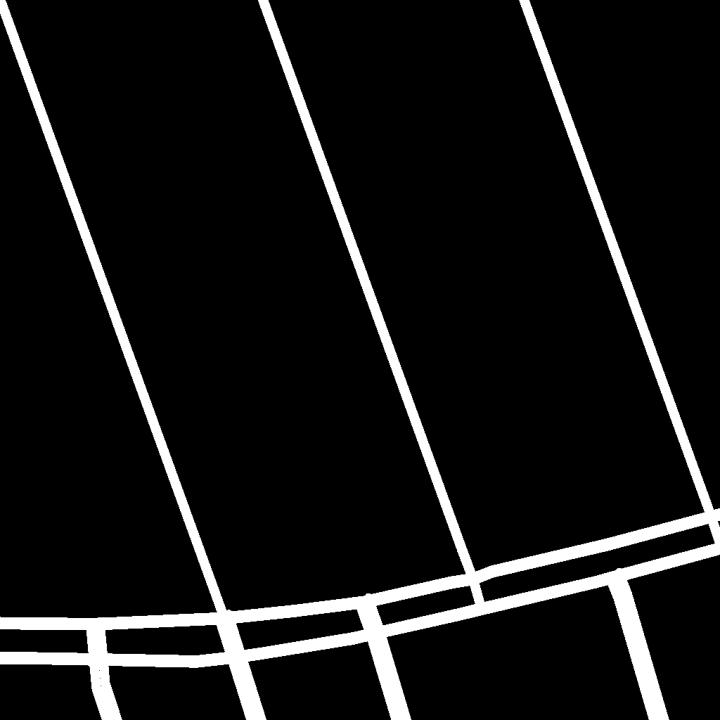} &
        \includegraphics[width=.194\linewidth]{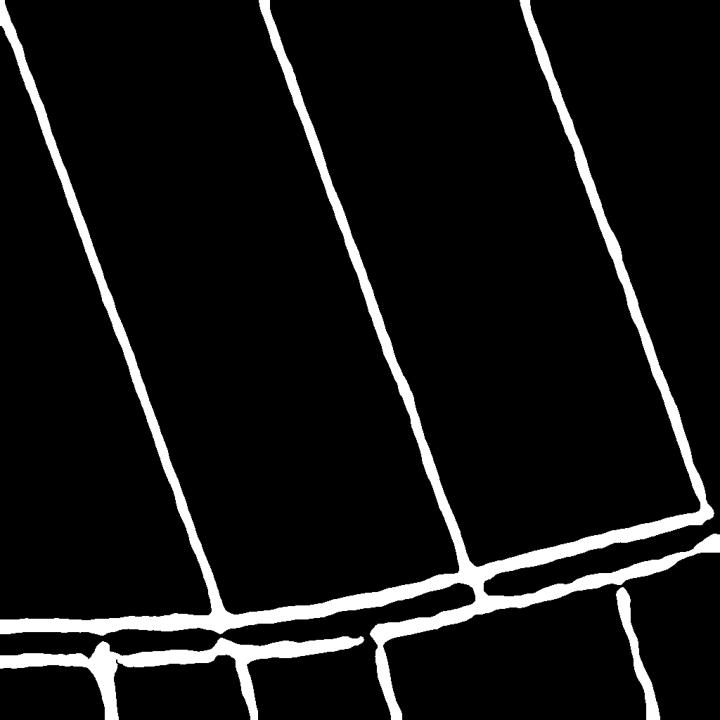} &
        \includegraphics[width=.194\linewidth]{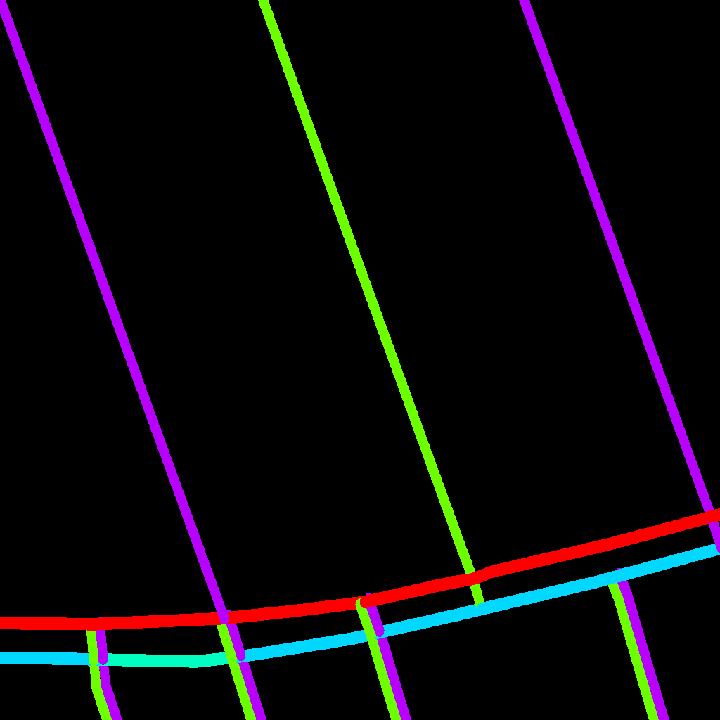} &
        \includegraphics[width=.194\linewidth]{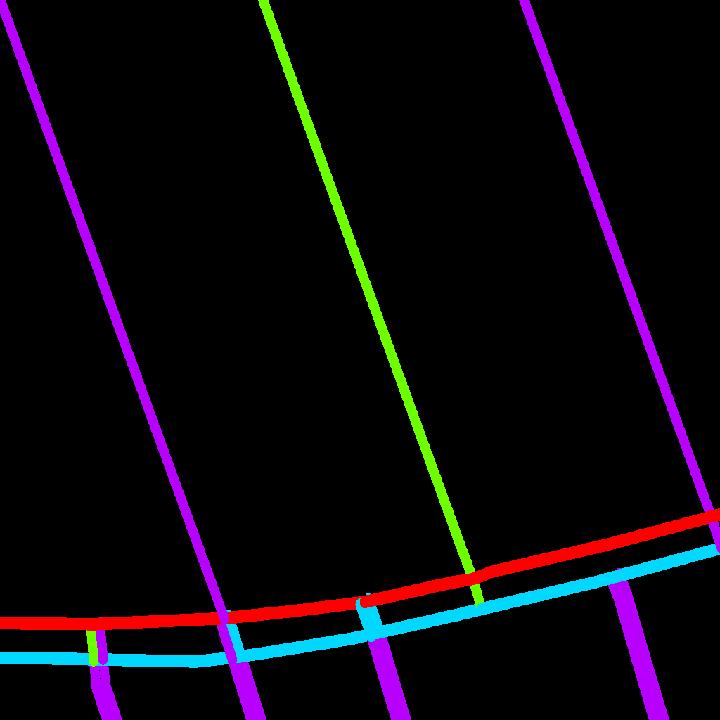} \\

        \includegraphics[width=.194\linewidth]{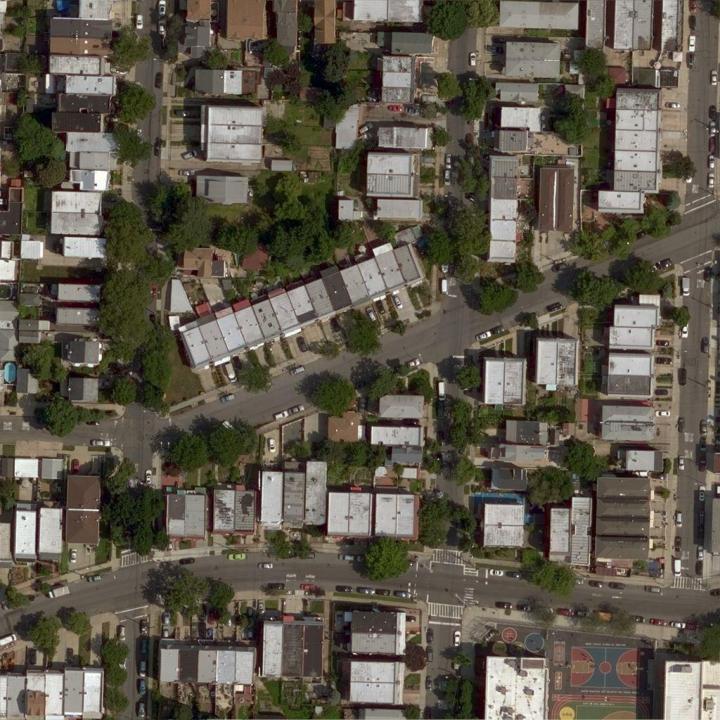} &
        \includegraphics[width=.194\linewidth]{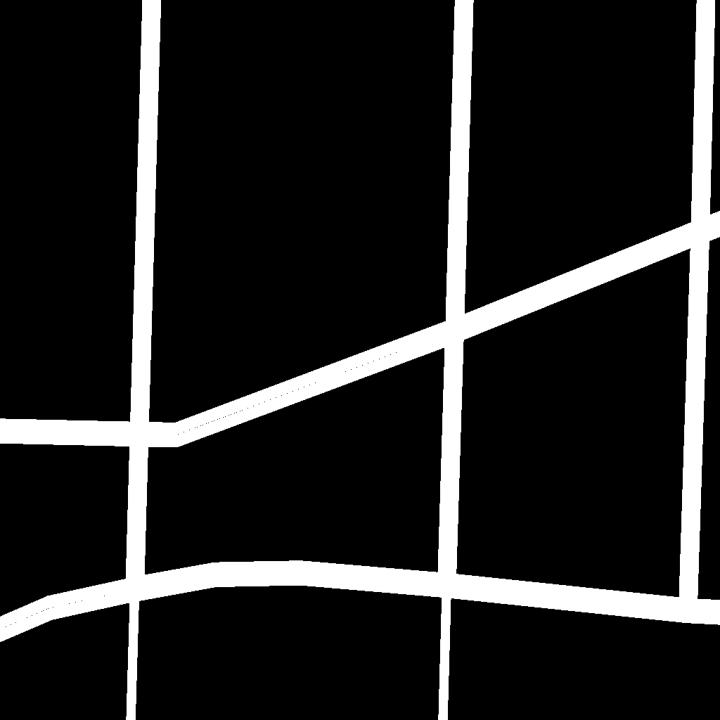} &
        \includegraphics[width=.194\linewidth]{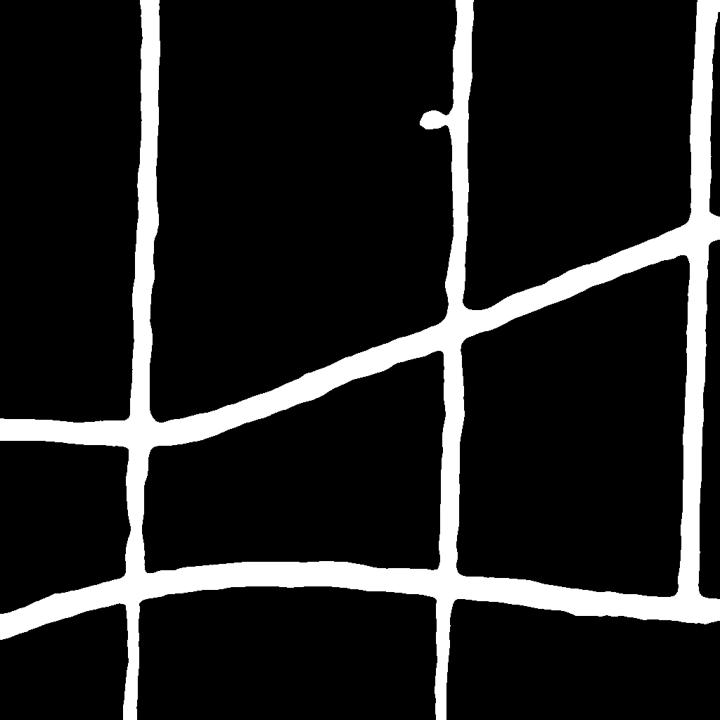} &
        \includegraphics[width=.194\linewidth]{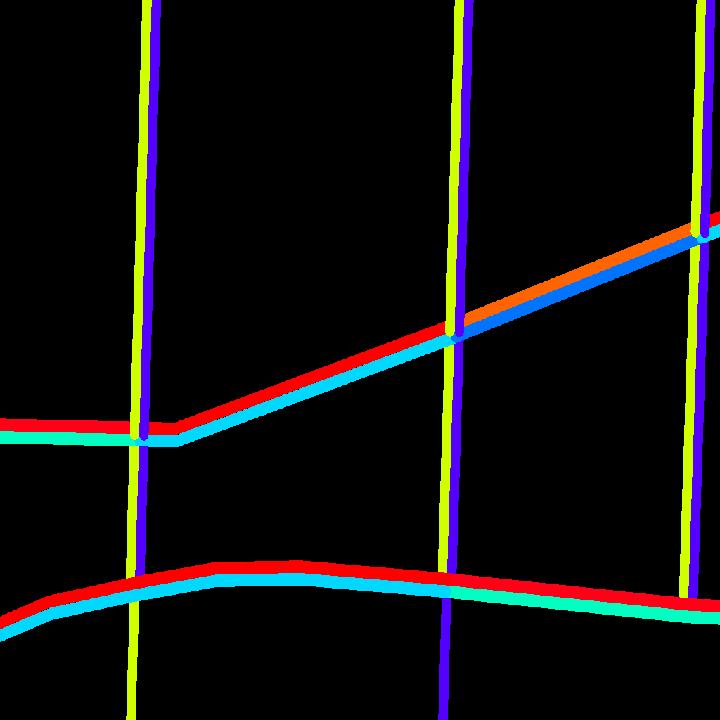} &
        \includegraphics[width=.194\linewidth]{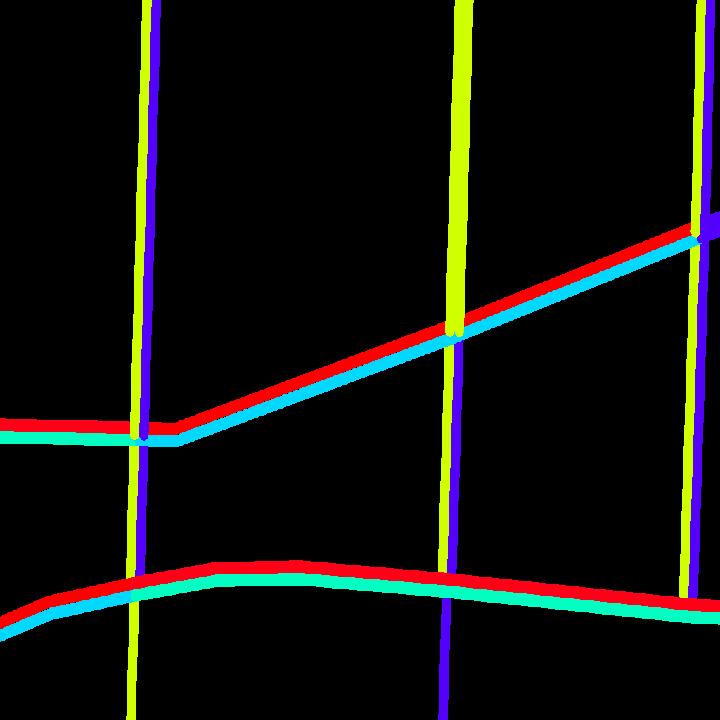} \\

        \includegraphics[width=.194\linewidth]{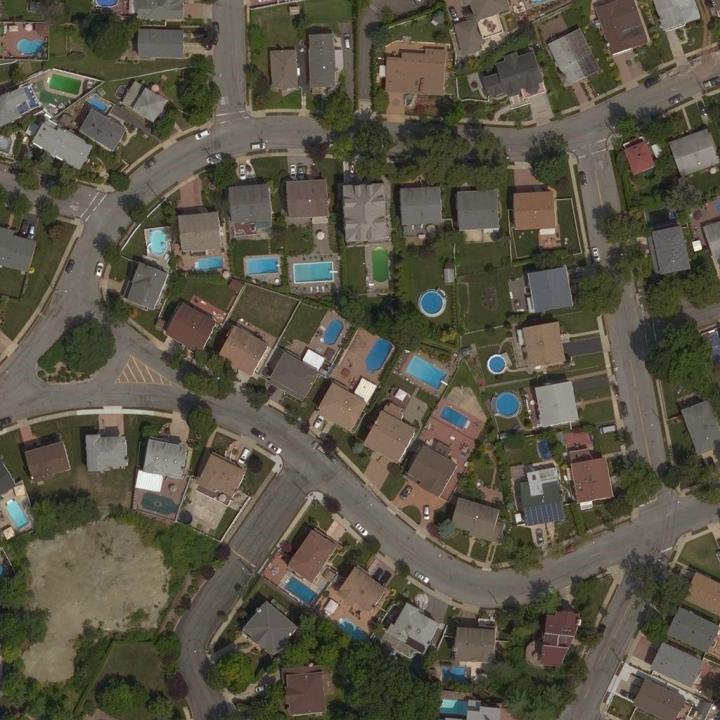} &
        \includegraphics[width=.194\linewidth]{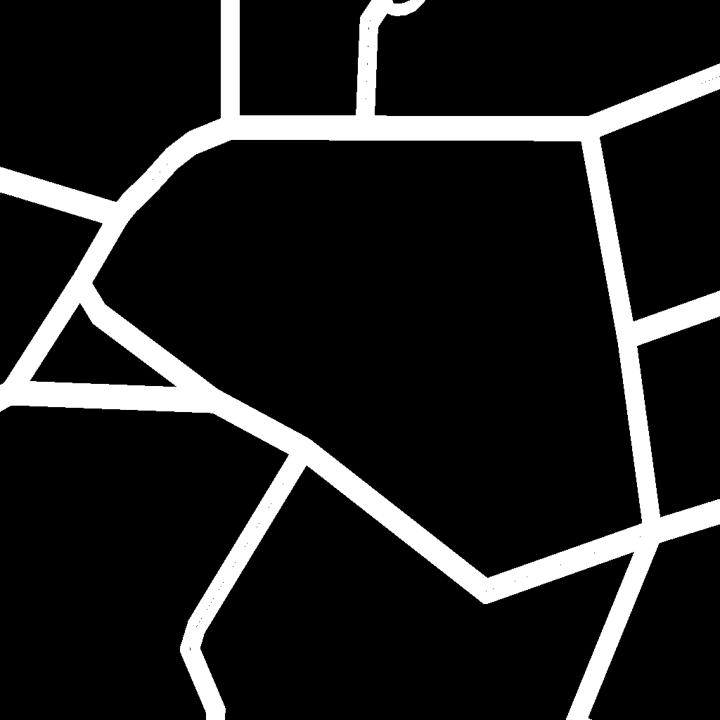} &
        \includegraphics[width=.194\linewidth]{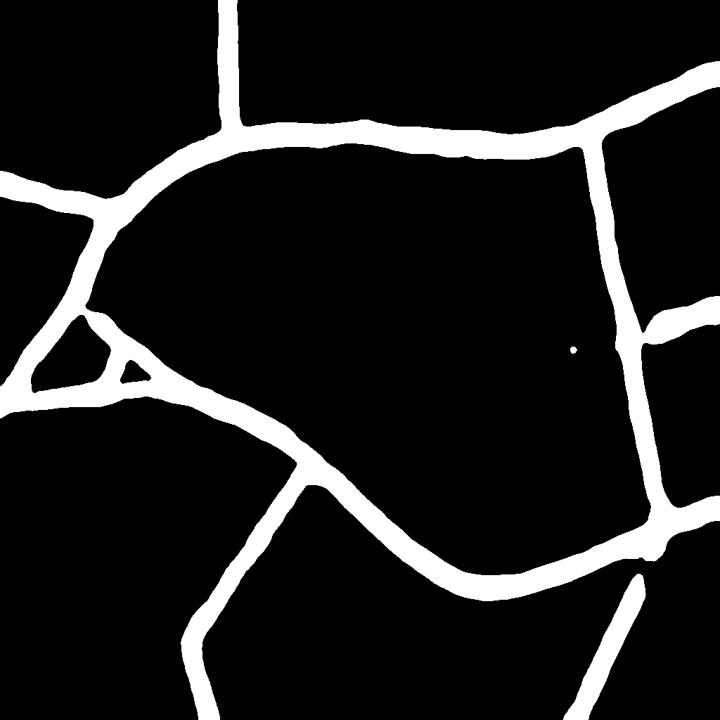} &
        \includegraphics[width=.194\linewidth]{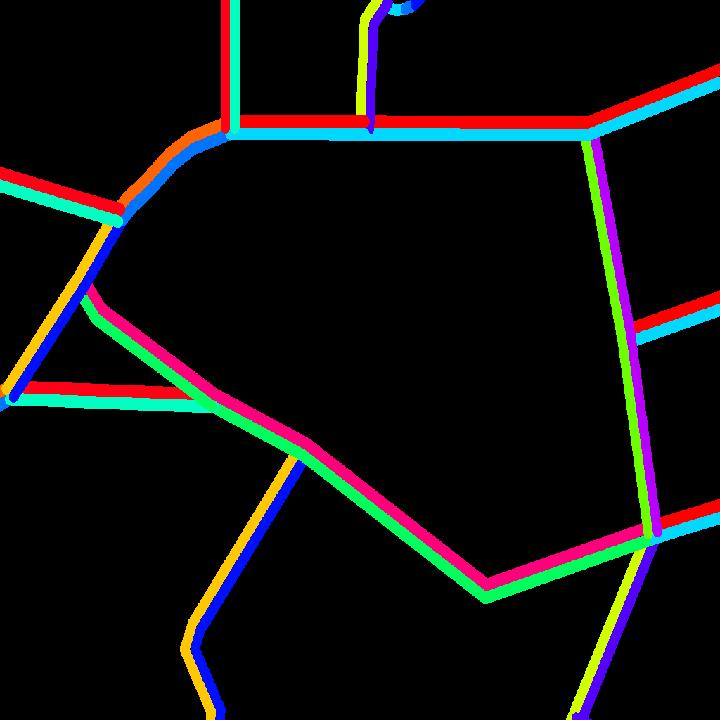} &
        \includegraphics[width=.194\linewidth]{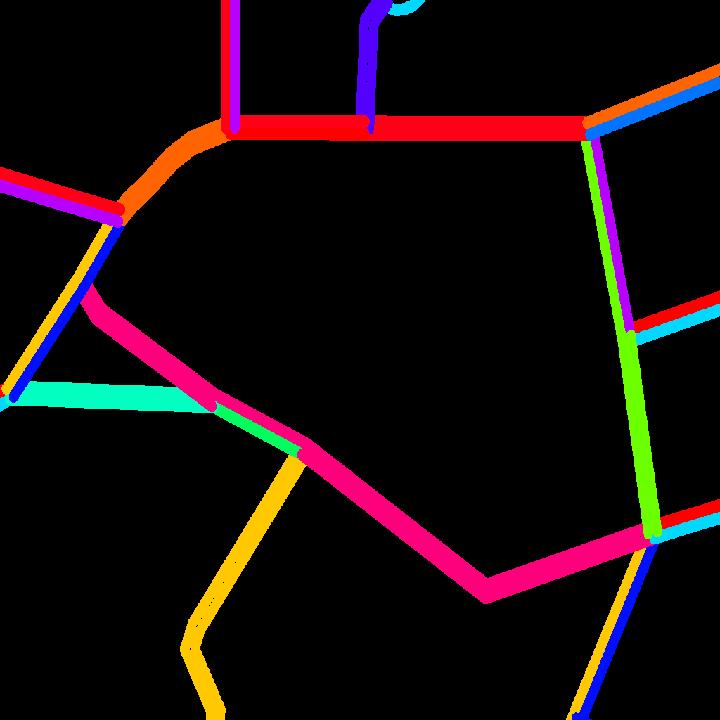} \\
    \end{tabular}

    \caption{Qualitative results for the auxiliary tasks of road segmentation and orientation estimation. Orientation is represented by $K=16$ bins using the HSV color map.}
    \label{fig:s_qualitative_aux}
\end{figure}

\begin{figure}
    \centering
    \includegraphics[width=.9\linewidth]{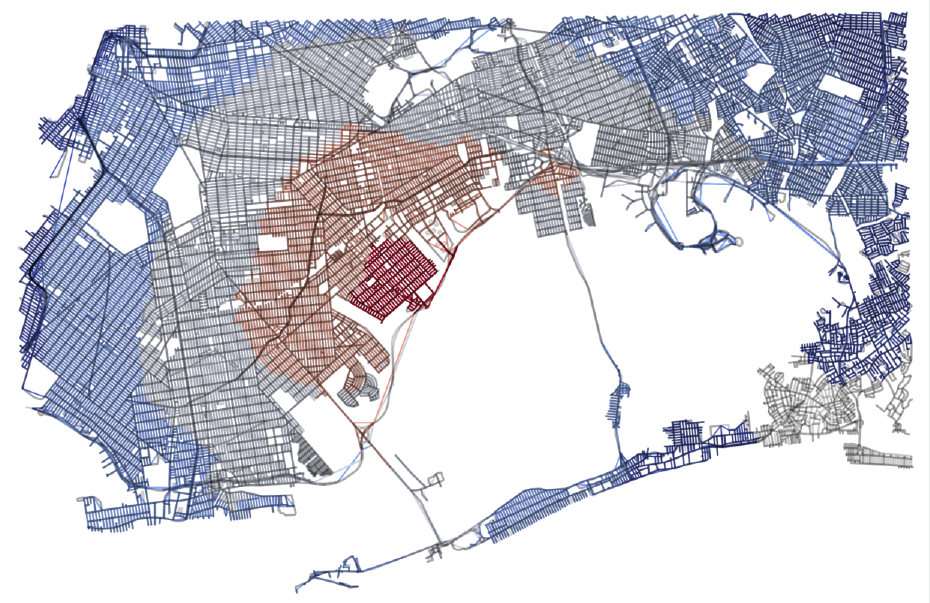}
    \caption{The output of our approach can be used for various applications, such as generating isochrone maps reflecting travel time. This map reflects travel times on a Monday at 8am. Starting from the center, each band reflects areas reachable in five minute increments (dark red shortest, dark blue furthest).}
    \label{fig:s_isochrone}
\end{figure}

\subsection{Auxiliary Tasks}

\figref{s_qualitative_aux} shows example outputs for road segmentation and orientation estimation. 
Given an overhead image, our method is able to identify roads and directions of travel. Combining these outputs with our estimates of traffic speeds enables our approach to estimate local motion models that describe traffic patterns. Finally, \tabref{s_results_aux} shows quantitative results for the auxiliary tasks. Though our objective was not necessarily to produce the best road detector or orientation estimator, these results show that our method is capable of capturing a notion of traversability.

\begin{table}
  \centering
  \caption{Evaluation of auxiliary tasks.}

  \setlength\tabcolsep{6pt}
  \begin{tabular}{@{}c|cc@{}}
    \toprule
    Speed & Road & Orientation \\
    (RMSE $\downarrow$) & (F1 Score $\uparrow$) & (Accuracy $\uparrow$) \\
    \bottomrule
    8.84 & 77.51\% & 73.78\% \\
    \bottomrule
  \end{tabular}
  
  \label{tbl:s_results_aux}
\end{table}

\subsection{Complexity Analysis}
Our full method has approximately 18.1 million trainable parameters and the total estimated model parameters size is 72.57 MB. It takes about 48 hours to train on a single NVIDIA A6000 (for fair comparison with baselines, we use a batch size of two and accumulate gradients across eight batches). We estimate that our approach has 1587.21 GFLOPS (floating point operations) and 792.41 GMACs (multiply-add operations). During inference, our unoptimized implementation achieves approximately 5.2 frames per second (batch size of 1).

\section{Application: Estimating Travel Time}

Our approach can be used to generate dense city-scale traffic models (in space and time). One potential application of this is generating travel time maps. We demonstrate this using OSMnx~\cite{boeing2017osmnx}, a library for visualizing street networks from OpenStreetMap. We correlate the output from our approach with the graph topology for New York City from OSMnx, using road segment lengths and our estimated speeds to assign travel times to each edge. \figref{s_isochrone} shows one such result, visualized as an isochrone map highlighting regions with similar travel time.

\section{Detailed Architecture}

We provide detailed architecture descriptions for the components of our network. \tabref{s_arch_encoder} shows the feature encoder. \tabref{s_arch_decoder} shows our MLP decoder used for generating the segmentation output. Finally, \tabref{s_arch_gtpe} shows the geo-temporal positional encoding module.

\begin{table}
  \centering
  \caption{Encoder architecture.}

  \setlength\tabcolsep{2pt}
  \resizebox{.9\linewidth}{!}{
      \begin{tabular}{@{}lcccc@{}}
        \toprule
        Layer (type:depth-idx) & Input Shape & Kernel Shape & Output Shape & Param \# \\
        \bottomrule
        Encoder & [1, 3, 1024, 1024] & -- & [1, 64, 256, 256] & -- \\
        — Sequential: 1-1 & [1, 3, 1024, 1024] & -- & [1, 64, 512, 512] & -- \\
        —    — Conv2d: 2-1 & [1, 3, 1024, 1024] & [3, 3] & [1, 48, 512, 512] & 1,296 \\
        —    — BatchNorm2d: 2-2 & [1, 48, 512, 512] & -- & [1, 48, 512, 512] & 96 \\
        —    — ReLU: 2-3 & [1, 48, 512, 512] & -- & [1, 48, 512, 512] & -- \\
        —    — Conv2d: 2-4 & [1, 48, 512, 512] & [3, 3] & [1, 64, 512, 512] & 27,648 \\
        —    — BatchNorm2d: 2-5 & [1, 64, 512, 512] & -- & [1, 64, 512, 512] & 128 \\
        —    — ReLU: 2-6 & [1, 64, 512, 512] & -- & [1, 64, 512, 512] & -- \\
        —    — Conv2d: 2-7 & [1, 64, 512, 512] & [3, 3] & [1, 64, 512, 512] & 36,864 \\
        — BatchNorm2d: 1-2 & [1, 64, 512, 512] & -- & [1, 64, 512, 512] & 128 \\
        — ReLU: 1-3 & [1, 64, 512, 512] & -- & [1, 64, 512, 512] & -- \\
        — MaxPool2d: 1-4 & [1, 64, 512, 512] & 3 & [1, 64, 256, 256] & -- \\
        — ModuleList: 1-5 & -- & -- & -- & -- \\
        —    — MBConvBlock: 2-8 & [1, 64, 256, 256] & -- & [1, 64, 256, 256] & 44,688 \\
        —    — MBConvBlock: 2-9 & [1, 64, 256, 256] & -- & [1, 64, 256, 256] & 44,688 \\
        — ModuleList: 1-6 & -- & -- & -- & -- \\
        —    — MBConvBlock: 2-10 & [1, 64, 256, 256] & -- & [1, 128, 128, 128] & 61,200 \\
        —    — MBConvBlock: 2-11 & [1, 128, 128, 128] & -- & [1, 128, 128, 128] & 171,296 \\
        —    — MBConvBlock: 2-12 & [1, 128, 128, 128] & -- & [1, 128, 128, 128] & 171,296 \\
        — ContextEncoder: 1-7 & [1, 2, 1024, 1024] & -- & [1, 64, 1024, 1024] & -- \\
        —    — LocationEncoder: 2-13 & [1, 2, 1024, 1024] & -- & [1, 64, 1024, 1024] & 12,736 \\
        —    — TimeEncoder: 2-14 & [1, 4] & -- & [1, 64] & 12,800 \\
        —    — LocationTimeEncoder: 2-15 & [1, 2, 1024, 1024] & -- & [1, 64, 1024, 1024] & 42,304 \\
        — Conv2d: 1-8 & [1, 64, 1024, 1024] & [3, 3] & [1, 256, 1024, 1024] & 147,712 \\
        — ModuleList: 1-9 & -- & -- & -- & -- \\
        —    — MHSABlock: 2-16 & [1, 128, 128, 128] & -- & [1, 4096, 256] & 1,213,704 \\
        —    — MHSABlock: 2-17 & [1, 4096, 256] & -- & [1, 4096, 256] & 918,024 \\
        —    — MHSABlock: 2-18 & [1, 4096, 256] & -- & [1, 4096, 256] & 918,024 \\
        —    — MHSABlock: 2-19 & [1, 4096, 256] & -- & [1, 4096, 256] & 918,024 \\
        —    — MHSABlock: 2-20 & [1, 4096, 256] & -- & [1, 4096, 256] & 918,024 \\
        — Conv2d: 1-10 & [1, 64, 1024, 1024] & [3, 3] & [1, 512, 1024, 1024] & 295,424 \\
        — ModuleList: 1-11 & -- & -- & -- & -- \\
        —    — MHSABlock: 2-21 & [1, 256, 64, 64] & -- & [1, 1024, 512] & 4,363,784 \\
        —    — MHSABlock: 2-22 & [1, 1024, 512] & -- & [1, 1024, 512] & 3,182,600 \\    
        \bottomrule
      \end{tabular}
  }
  
  \label{tbl:s_arch_encoder}
\end{table}

\begin{table}
  \centering
  \caption{Decoder architecture.}

  \setlength\tabcolsep{2pt}
  \resizebox{.8\linewidth}{!}{
      \begin{tabular}{@{}lcccc@{}}
        \toprule
        Layer (type:depth-idx) & Input Shape & Kernel Shape & Output Shape & Param \# \\
        \bottomrule
        Decoder & [1, 64, 256, 256] & -- & [1, 1, 1024, 1024] & -- \\
        — MLP: 1-1 & [1, 512, 32, 32] & -- & [1, 1024, 512] & -- \\
        —    — Linear: 2-1 & [1, 1024, 512] & -- & [1, 1024, 512] & 262,656 \\
        — MLP: 1-2 & [1, 256, 64, 64] & -- & [1, 4096, 512] & -- \\
        —    — Linear: 2-2 & [1, 4096, 256] & -- & [1, 4096, 512] & 131,584 \\
        — MLP: 1-3 & [1, 128, 128, 128] & -- & [1, 16384, 512] & -- \\
        —    — Linear: 2-3 & [1, 16384, 128] & -- & [1, 16384, 512] & 66,048 \\
        — MLP: 1-4 & [1, 64, 256, 256] & -- & [1, 65536, 512] & -- \\
        —    — Linear: 2-4 & [1, 65536, 64] & -- & [1, 65536, 512] & 33,280 \\
        — Sequential: 1-5 & [1, 2048, 256, 256] & -- & [1, 512, 256, 256] & -- \\
        —    — Conv2d: 2-5 & [1, 2048, 256, 256] & [1, 1] & [1, 512, 256, 256] & 1,048,576 \\
        —    — BatchNorm2d: 2-6 & [1, 512, 256, 256] & -- & [1, 512, 256, 256] & 1,024 \\
        —    — ReLU: 2-7 & [1, 512, 256, 256] & -- & [1, 512, 256, 256] & -- \\
        — Dropout2d: 1-6 & [1, 512, 256, 256] & -- & [1, 512, 256, 256] & -- \\
        — Conv2d: 1-7 & [1, 512, 256, 256] & [1, 1] & [1, 1, 256, 256] & 513 \\
        \bottomrule
      \end{tabular}
  }
  
  \label{tbl:s_arch_decoder}
\end{table}

\begin{table}
  \centering
  \caption{Geo-temporal positional encoding (GTPE) module.}

  \setlength\tabcolsep{2pt}
  \resizebox{.8\linewidth}{!}{
      \begin{tabular}{@{}lcccc@{}}
        \toprule
        Layer (type:depth-idx) & Input Shape & Output Shape & Param \# \\
        \bottomrule
        GTPE & [1, 2, 1024, 1024] & [1, 64, 1024, 1024] & -- \\
        — LocationEncoder: 1-1 & [1, 2, 1024, 1024] & [1, 64, 1024, 1024] & -- \\
        —    — LocParam: 2-1 & [1, 2, 1024, 1024] & [1, 3, 1024, 1024] & -- \\
        —    — SirenNet: 2-2 & [1048576, 3] & [1048576, 64] & -- \\
        —    —    — ModuleList: 3-1 & -- & -- & -- \\
        —    —    —    — Siren: 4-1 & [1048576, 3] & [1048576, 64] & 256 \\
        —    —    —    —    — Linear: 5-1 & [1048576, 3] & [1048576, 64] & 256 \\
        —    —    —    —    — Sine: 5-2 & [1048576, 64] & [1048576, 64] & -- \\
        —    —    —    — Siren: 4-2 & [1048576, 64] & [1048576, 64] & 4,160 \\
        —    —    —    —    — Linear: 5-3 & [1048576, 64] & [1048576, 64] & 4,160 \\
        —    —    —    —    — Sine: 5-4 & [1048576, 64] & [1048576, 64] & -- \\
        —    —    —    — Siren: 4-3 & [1048576, 64] & [1048576, 64] & 4,160 \\
        —    —    —    —    — Linear: 5-5 & [1048576, 64] & [1048576, 64] & 4,160 \\
        —    —    —    —    — Sine: 5-6 & [1048576, 64] & [1048576, 64] & -- \\
        —    —    — Siren: 3-2 & [1048576, 64] & [1048576, 64] & 4,160 \\
        —    —    —    — Linear: 4-4 & [1048576, 64] & [1048576, 64] & 4,160 \\
        —    —    —    — Identity: 4-5 & [1048576, 64] & [1048576, 64] & -- \\
        — TimeEncoder: 1-2 & [1, 4] & [1, 64] & -- \\
        —    — TimeParam: 2-3 & [1, 4] & [1, 4] & -- \\
        —    — SirenNet: 2-4 & [1, 4] & [1, 64] & -- \\
        —    —    — ModuleList: 3-3 & -- & -- & -- \\
        —    —    —    — Siren: 4-6 & [1, 4] & [1, 64] & 320 \\
        —    —    —    —    — Linear: 5-7 & [1, 4] & [1, 64] & 320 \\
        —    —    —    —    — Sine: 5-8 & [1, 64] & [1, 64] & -- \\
        —    —    —    — Siren: 4-7 & [1, 64] & [1, 64] & 4,160 \\
        —    —    —    —    — Linear: 5-9 & [1, 64] & [1, 64] & 4,160 \\
        —    —    —    —    — Sine: 5-10 & [1, 64] & [1, 64] & -- \\
        —    —    —    — Siren: 4-8 & [1, 64] & [1, 64] & 4,160 \\
        —    —    —    —    — Linear: 5-11 & [1, 64] & [1, 64] & 4,160 \\
        —    —    —    —    — Sine: 5-12 & [1, 64] & [1, 64] & -- \\
        —    —    — Siren: 3-4 & [1, 64] & [1, 64] & 4,160 \\
        —    —    —    — Linear: 4-9 & [1, 64] & [1, 64] & 4,160 \\
        —    —    —    — Identity: 4-10 & [1, 64] & [1, 64] & -- \\
        — LocationTimeEncoder: 1-3 & [1, 2, 1024, 1024] & [1, 64, 1024, 1024] & -- \\
        —    — LocParam: 2-5 & [1, 2, 1024, 1024] & [1, 3, 1024, 1024] & -- \\
        —    — TimeParam: 2-6 & [1, 4] & [1, 4] & -- \\
        —    — SirenNet: 2-7 & [1048576, 7] & [1048576, 64] & -- \\
        —    —    — ModuleList: 3-5 & -- & -- & -- \\
        —    —    —    — Siren: 4-11 & [1048576, 7] & [1048576, 128] & 1,024 \\
        —    —    —    —    — Linear: 5-13 & [1048576, 7] & [1048576, 128] & 1,024 \\
        —    —    —    —    — Sine: 5-14 & [1048576, 128] & [1048576, 128] & -- \\
        —    —    —    — Siren: 4-12 & [1048576, 128] & [1048576, 128] & 16,512 \\
        —    —    —    —    — Linear: 5-15 & [1048576, 128] & [1048576, 128] & 16,512 \\
        —    —    —    —    — Sine: 5-16 & [1048576, 128] & [1048576, 128] & -- \\
        —    —    —    — Siren: 4-13 & [1048576, 128] & [1048576, 128] & 16,512 \\
        —    —    —    —    — Linear: 5-17 & [1048576, 128] & [1048576, 128] & 16,512 \\
        —    —    —    —    — Sine: 5-18 & [1048576, 128] & [1048576, 128] & -- \\
        —    —    — Siren: 3-6 & [1048576, 128] & [1048576, 64] & 8,256 \\
        —    —    —    — Linear: 4-14 & [1048576, 128] & [1048576, 64] & 8,256 \\
        —    —    —    — Identity: 4-15 & [1048576, 64] & [1048576, 64] & -- \\
      \end{tabular}
  }
  
  \label{tbl:s_arch_gtpe}
\end{table}

\end{document}